\documentclass[10pt,twocolumn,letterpaper]{article}
\usepackage{titling}

\usepackage[font=small]{caption}
\usepackage{bm}
\usepackage{iccv}
\usepackage{times}
\usepackage{epsfig}
\usepackage{graphicx}
\usepackage{amsmath}
\usepackage{amssymb}
\usepackage{lib}

\usepackage{booktabs}
\usepackage[table]{xcolor}
\usepackage{bm}
\usepackage{stmaryrd}
\usepackage{multirow}
\usepackage[accsupp]{axessibility} 
\usepackage{subcaption}
\usepackage{color}
\usepackage{cuted}
\usepackage{etoolbox}
\usepackage[nocompress]{cite}

\AfterEndEnvironment{strip}

\usepackage[pagebackref,breaklinks,colorlinks]{hyperref}

\usepackage{cuted}

\iccvfinalcopy 

\def\iccvPaperID{7185} 

\newcommand{\ours}[0]{DiffHOI} 

\begin{document}

\title{Diffusion-Guided Reconstruction of Everyday Hand-Object Interaction Clips}

\author{Yufei Ye \qquad Poorvi Hebbar \qquad Abhinav Gupta \qquad Shubham Tulsiani  \\
Carnegie Mellon University  \\
{\tt \small \{yufeiy2, phebbar, gabhinav, shubhtuls\}@andrew.cmu.edu} \\
{\tt \small \href{https://judyye.github.io/diffhoi-www}{https://judyye.github.io/diffhoi-www}}
}


\maketitle

\begin{strip}
\centering
\vspace{-1.2cm}
\includegraphics[width=\textwidth]{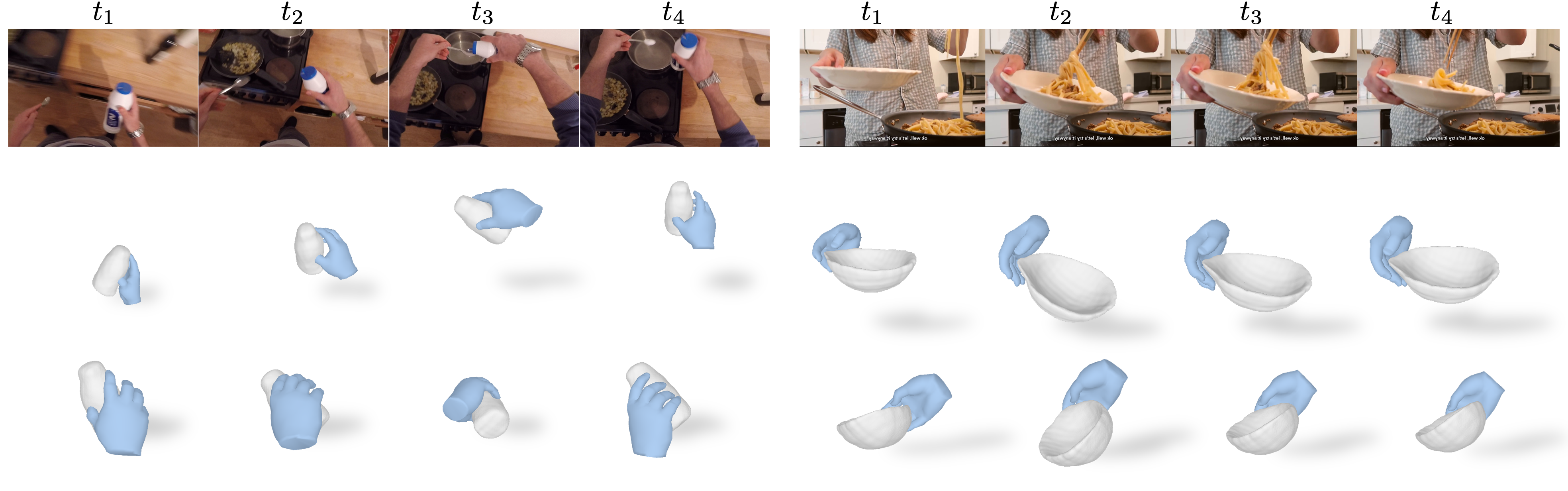}






\captionof{figure}{
Given a video clip depicting a hand-object interaction, we infer the underlying 3D shape of both the hand and the object. \textbf{Top:} sampled input frames; \textbf{Middle:} reconstruction visualized in the image frame; \textbf{Bottom:} reconstruction from a novel view. Please see the website for reconstruction videos. 
}
\label{fig:teaser}
\end{strip}

\ificcvfinal\thispagestyle{empty}\fi

\begin{abstract}
We tackle the task of reconstructing  hand-object interactions from short video clips. Given an input video, our approach casts 3D inference as a per-video optimization and recovers a neural 3D representation of the object shape, as well as the time-varying motion and hand articulation. While the input video naturally provides some multi-view cues to guide 3D inference, these are insufficient on their own due to occlusions and limited viewpoint variations. To obtain accurate 3D, we augment the multi-view signals with generic data-driven priors to guide reconstruction. Specifically, we learn a diffusion network to model the conditional distribution of (geometric) renderings of objects conditioned on hand configuration and category label, and leverage it as a prior to guide the novel-view renderings of the reconstructed scene. We empirically evaluate our approach on egocentric videos across 6 object categories, and observe significant improvements over prior single-view and multi-view methods. Finally, we demonstrate our system's ability to reconstruct arbitrary clips from YouTube, showing both $1^{st}$ and $3^{rd}$ person interactions.
\end{abstract}

\section{Introduction}


Our hands allow us to affect the world around us. From pouring the morning coffee to clearing the dinner table, we continually use our hands to interact with surrounding objects. In this work, we pursue the task of understanding such everyday interactions in 3D. Specifically, given a short clip of a human interacting with a rigid object, our approach can infer the shape of the underlying object as well as its (time-varying) relative transformation \wrt an articulated hand (see Fig.~\ref{fig:teaser} for sample results).

This task of recovering 3D representations of hand-object interactions (HOI) has received growing interest. While initial approaches~\cite{Hasson2020LeveragingPC,tekin2019h+,liu2021semi,garcia2018first,Cao2020ReconstructingHI} framed it as 6-DoF pose task estimation for known 3D objects/templates, subsequent methods have tackled the reconstruction of apriori unknown objects~\cite{ye2022hand,karunratanakul2020grasping,hasson19_obman}. Although single-view 3D reconstruction approaches can leverage data-driven techniques to reconstruct HOI images~\cite{hasson19_obman, alli2015TheYO,ye2022hand}, 
these approaches cannot obtain precise reconstructions given the fundamentally limited nature of the single-view input. On the other hand, current video-based HOI reconstruction methods primarily exploit multi-view cues and rely on purely geometry-driven optimization for reconstruction. As a result, these methods are suited for in-hand scanning where a user carefully presents exhaustive views of the object of interest, but they are not applicable to our setting as aspects of the object may typically be unobserved.




Towards enabling accurate reconstruction given short everyday interaction clips, our approach (\ours) unifies the data-driven and the geometry-driven techniques. Akin to the prior video-based reconstruction methods, we frame the reconstruction task as that of optimizing a video-specific temporal scene representation. However, instead of purely relying on geometric reprojection errors, we also incorporate data-driven priors to guide the optimization. In particular, we learn a 2D diffusion network which models the distribution over plausible (geometric) object renderings conditioned on estimated hand configurations.  Inspired by  recent applications in text-based 3D generation~\cite{poole2022dreamfusion,lin2022magic3d}, we use this diffusion model as a generic data-driven regularizer for the video-specific 3D optimization.

We empirically evaluate our system across several first-person hand-object interaction clips from the HOI4D dataset~\cite{hoi4d}, and show that it significantly improves over both prior single-view and multi-view methods.  To demonstrate its applicability in more general settings, we also show qualitative results on arbitrary interaction clips from YouTube, including both first-person and third-person clips. 



\section{Related Works}




\paragraph{Reconstructing Hand-Object Interactions. } Hands and objects  inherently undergo mutual occlusions which makes 3D reconstruction extremely ill-posed during interactions. 
Hence, many works~\cite{hasson2021towards,rhoi2020,tekin2019h+, Patel2022,zhang2020phosa,Hasson2020LeveragingPC,oikonomidis2011full,tzionas2016capturing,fan2023arctic,bhatnagar2022behave} reduce the problem to 6D pose estimation by assuming access to instance-specific templates. Their frameworks of 6D pose optimization can be applied to both videos and images. Meanwhile, template-free methods follow two paradigms for videos and images. On one hand, video-based methods takes in synchronized RGB(D) videos~\cite{jiang2022neuralhofusion,guo2019relightables,sun2021neural,suo2021neuralhumanfvv,zhang2023neuraldome} or monocular videos~\cite{huang2022hhor,hampali2023hand,wen2023bundlesdf} and fuse observation to a canonical 3D representation~\cite{newcombe2015dynamicfusion}. This paradigm does not use any prior knowledge and requires all regions being observed in some frames, which is often not true in everyday video clips . On the other hand, methods that reconstruct HOI from single images~\cite{ye2022hand,hasson19_obman,chen2022alignsdf,chen2023gsdf,karunratanakul2020grasping} leverage learning-based prior to reconstruct more general objects. While they are able to generate reasonable per-frame predictions, it is not trivial to aggregate information from multiple views in one sequence and generate a time-consistent 3D shape. Our work is template-free and unifies both geometry-driven and data-driven methods.

\vspace{-0.3cm}\paragraph{Generating Hand-Object Interactions.} Besides reconstructing the ongoing HOIs, many works have explored generating plausible HOIs in different formulations. Some works model their joint distributions~\cite{taheri2022goal,karunratanakul2020grasping,hu2022hand,cao2021reconstructing}.  Works that are usually called afforadance or grasp prediction study the conditional distribution that generates hands/humans given an object/scenes, in 3D representation~\cite{li2019putting,jiang2021hand,corona2020ganhand,brahmbhatt2020contactpose,taheri2020grab} or 2D images~\cite{ye2023affordance,kulal2023putting,chuang2018learning,wang2017binge}. Recently, some other works explore the reverse problem that generates plausible scenes for a given human/hand pose~\cite{YangCVPR2022OakInk,ye2022scene,petrov2023object,nie2022pose2room,zheng2021inferring,brooks2022hallucinating}. 
In our work, we follow the latter formulation to learn an image-based generative model of hand-held objects given hands, since hand pose estimation~\cite{rong2021frankmocap} is good enough to bootstrap the system.

\vspace{-0.3cm}\paragraph{Neural Implicit Fields for Dynamic Scenes. }
Neural implicit fields \cite{park2019deepsdf, mescheder2019occupancy,mildenhall2021nerf} are {flexible representation that allows capturing diverse shape with various topology and can be optimized with 2D supervision via differentiable rendering }\cite{yariv2021volume,wang2021neus}. To extend them to dynamic scenes, a line of work optimizes per-scene representation with additional general motion priors \cite{park2021hypernerf,pumarola2021d,park2021nerfies,li2023dynibar,lu2021omnimatte}. Though these methods can synthesize highly realistic novel views from nearby angles, they struggle to extrapolate viewpoints \cite{gao2022monocular}. Another line of work incorporate category-specific priors \cite{yang2022banmo,tulsiani2020implicit,wu2022magicpony,mu2021sdf,wei2022self} to model articulations. They typically work on single deformable objects of the same category such as quadrupeds or drawers.  In contrast, we focus on hand-object interactions across six rigid categories where dynamics are mostly due to changing spatial relations and hand articulations.

\vspace{-0.3cm}\paragraph{Distilling diffusion models. }
Diffusion models~\cite{Ho2020DenoisingDP, Rombach2021HighResolutionIS} have made significant strides in text-to-image synthesis. They can also be quickly adapted to take additional inputs or generate outputs in other domains~\cite{zhang2023adding,hu2021lora}.
Recent works have shown that these image-based diffusion models can be distilled~\cite{wang2023score,poole2022dreamfusion} into 3D scene representations for generation or reconstruction~\cite{jun2023shap, poole2022dreamfusion, zhou2023sparsefusion,liu2023zero,watson2022novel,melaskyriazi2023realfusion,Deng2022NeRDiSN}. Inspired by them, we also adopt a conditional diffusion model for guiding 3D inference of HOIs. However, instead of learning prior over appearance, we use a diffusion model to learn a prior over a-modal geometric renderings. 


\section{Method}

\begin{figure*}
    \centering
\includegraphics[width=\linewidth]{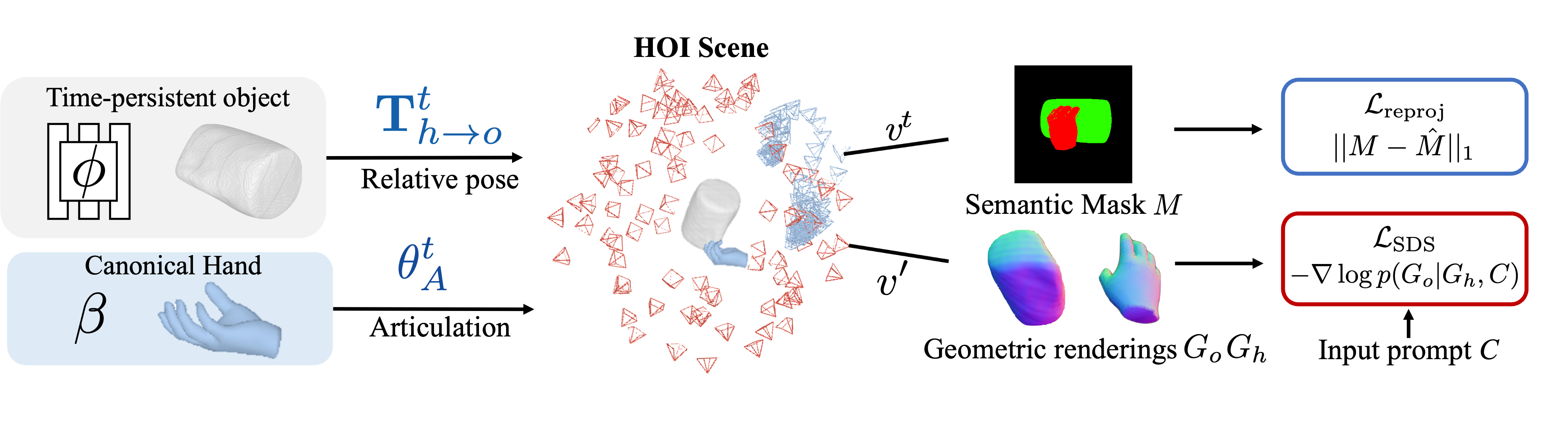}    
\vspace{-1em}
    \caption{\textbf{Method Overview:} We decompose the HOI scene into 1) a rigid  time-persistent implicit field $\phi$ for the object, 2) hand meshes parameterized by hand shape $\beta$ and articulations $\theta_A^t$, and 3)  their time-varying  relative poses $T_{h\to o}^t$. We define the camera poses $T^t_{c\to h}$ in the hand frame. The scene representation is optimized with respect to a reprojection term from the original views $v^t$ and a data-prior term from novel views $v'$.
    }
    \label{fig:overview}
\end{figure*}


Given a monocular video of a hand interacting with a rigid object, we aim to reconstruct the underlying hand-object interaction, \ie, the 3D shape of the   object, its pose in every frame, along with per-frame  hand meshes  and camera poses.  We frame the inference as per-video optimization of an underlying 3D representations. While the multiple frames allow leveraging multi-view cues, they are not sufficient as the object of interests is often partially visible in everyday video clips, due to limited viewpoints and mutual occlusion. 
Our key insight is to incorporate both view consistency across multiple frames and a data-driven prior of the  HOIs geometry. The learned interaction prior captures both category priors, \eg mugs are generally cylindrical,  and hand priors, \eg pinched fingers are likely to hold thin handles. We train a conditional diffusion model for the prior that guides the HOI to be reconstructed during per-video optimization. 

More specifically, given a monocular video $\hat I^t$ with corresponding hand and object masks $\hat M^t \equiv (\hat M_h^t, \hat M_o^t)$, we aim to optimize a HOI representation (Sec. \ref{sec:hybrid_rep})  that consists of a time-persistent implicit field $\bm\phi$ for the rigid object, a time-varying morphable mesh for the hand $H^t$, the relative transformation between hand and object $T^t_{h\to o}$, and time-varying camera poses $T^t_{c\to h}$. The optimization objective consists of two terms (Sec. \ref{sec:loss}): a reprojection error from the estimated original viewpoint and data-driven prior term  that encourages the object geometry to appear more plausible given category and hand information when looking from another viewpoint. 
The prior is implemented as a diffusion model conditioned on a text prompt $C$ about  the category  and renderings of the hand $\pi(H)$ with  geometry cues (Sec. \ref{sec:diffusion}). It denoises  the rendering of the object $\pi(O)$ and 
backpropagates the gradient to the 3D HOI representation by score distillation sampling (SDS)~\cite{poole2022dreamfusion}.

\subsection{HOI Scene Representation} 
\label{sec:hybrid_rep}

\paragraph{Implicit field for the object.} The rigid object is represented by a time-persistent implicit field $\bm\phi$ that can handle unknown topology and has shown promising results when optimizing for challenging shapes~\cite{yariv2021volume,wang2021neus,yang2022banmo}. For every point in the object frame, we use multi-layer perceptrons to predict the signed distance function (SDF) to the object surface, $s = \bm\phi(\bm X)$. 

\vspace{-0.3cm}\paragraph{Time-varying hand meshes.} We use a pre-defined parametric mesh model MANO~\cite{mano} to represent hands across frames. The mesh can be animated by low-dimensional parameters and thus can better capture more structured motions, \ie hand articulation. 
We obtain  hand meshes $H^t$ in a canonical hand wrist frame by rigging MANO with a 45-dim pose parameters $\bm\theta^t_A$ and 10-dim shape  parameters $\bm\beta$, \ie $H^t = \text{MANO}(\bm\theta_A^t, \bm\beta)$. The canonical wrist frame is invariant to wrist orientation and only captures finger articulations. 



\vspace{-0.3cm}\paragraph{Composing to a scene. } Given the time-persistent object representation $\bm\phi$ and a time-varying hand mesh $H^t$, we then compose them into a scene at time $t$ such that they can be reprojected back to the image space from the cameras. Prior works~\cite{hasson2021towards,Patel2022,hampali2020honnotate} typically track 6D object pose directly in the camera frame $T_{c\to o}$ which requires an object template to define the object pose.  In our case, since we do not have access to object templates, the object pose in the camera frame is hard to estimate directly. 
Instead, we track object pose with respect to hand wrist $T^t_{h\to o}$ and initialize them to identity. It is based on the observation that the object of interest usually moves together with the hand and undergoes ``common fate" \cite{cohesive}.
A point in the rigid object frame can be related to  the predicted camera frame by composing the two transformations, camera-to-hand $T^t_{c\to h}$ and hand-to-object $T^t_{h \to o}$. For notation convention, we denote the implicit field transformed to  the hand frame at time t as $\bm\phi^t(\cdot) \equiv \bm\phi(T_{h\to o} (\cdot))$. Besides modeling camera extrinsics, we also optimize for per-frame camera intrinsics $\bm K^t$ to account for zoom-in effect, cropping operation, and inaccurate intrinsic estimation. 

In summary, given a monocular video with corresponding masks, the parameters to be optimized are 
\begin{align}
\bm\phi, \bm\beta, \bm\theta^t_A, T^t_{h\to o}, T^t_{c\to h},  \bm K^t
\end{align}

\begin{figure}
    \centering
    \includegraphics[width=\linewidth]{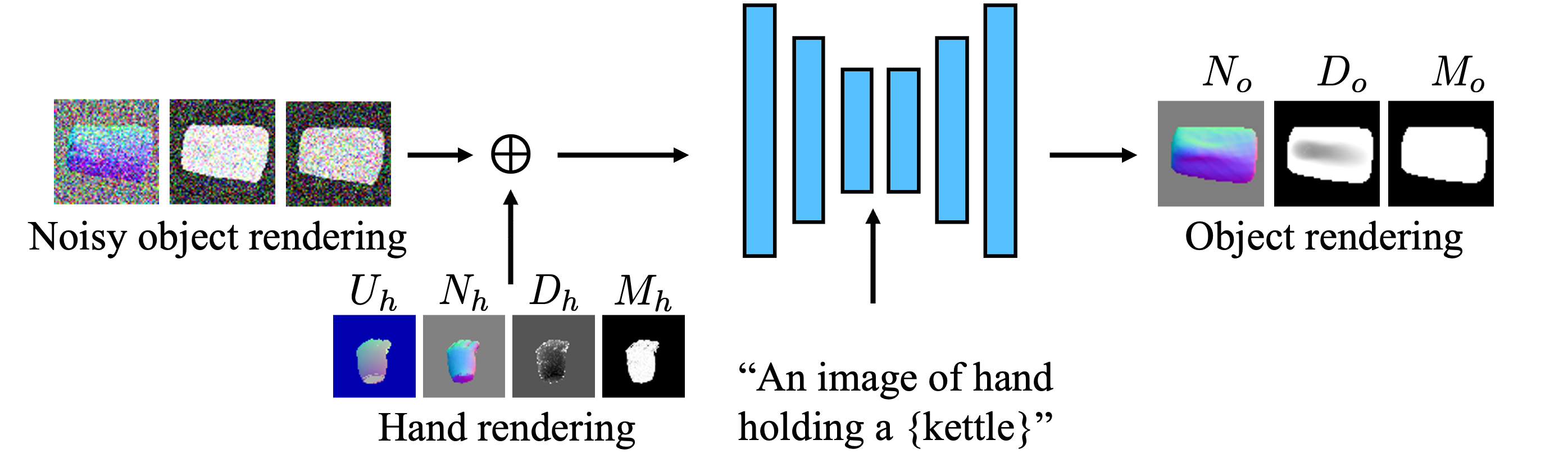}
    \caption{\textbf{Geometry-informed Diffusion Model:} The diffusion model takes in a noisy geometry rendering of the object, the geometry rendering of the hand, and a text prompt, to output the denoised geometry rendering of objects.}
    \label{fig:diffusion}
\end{figure}

\vspace{-0.3cm}\paragraph{Differentialble Rendering. } To render the HOI scene into an image, we separately render the object (using volumetric rendering\cite{yariv2021volume}) and the hand (using mesh rendering\cite{liu2019soft, paszke2019pytorch}) to obtain geometry cues.  We then blend their renderings into HOI images by their rendered depth. 

Given an arbitrary viewpoint $v$, both differentiable renders can render  geometry images  including mask, depth, and normal images, \ie $G_h \equiv (M_h, D_h, N_h), G_o \equiv (M_o, D_o, N_o)$ To compose them into a semantic mask $M_{HOI}$ that is later used to calculate the reprojection loss, we softly blend the individual masks by their predicted depth. Similar to blending two-layer surfaces of mesh rendering, the final semantic masks can be computed by alpha blending: $M = B(M_h, M_o, D_h, D_o)$. Please refer to supplementary material for the full derivation of the blending function $B$.

\subsection{Data-Driven Prior for Geometry}
\label{sec:diffusion}
When observing everyday interactions, we do not directly observe all aspects of the object because of occlusions and limited viewpoint variability. Despite this, we aim to reconstruct the 3D shape of the full object. To do so, we rely on a data-driven prior that captures the likelihood of a common object geometry given its category and the hand interacting with it $p(\bm\phi^t | H^t, C)$. More specifically, we use a diffusion model which learns a data-driven distribution over geometry rendering of objects given that of hands and category. 
\begin{align}
\log p(\bm\phi^t | H^t, C) \approx \mathbb E_{v\sim V} \log p(\pi(\bm\phi^t ; v) | \pi(H^t ; v), C)   
\end{align}
where $v\sim V$ is a viewpoint drawn from a prior distribution, $C$ as category label and $\pi$ as rendering function.
Since this learned prior only operates in geometry domain, there is no domain gap to transfer the prior across daily videos with complicated appearances.  We first pretrain this diffusion model with large-scale ground truth HOIs and then use the learned prior to guide per-sequence optimization (Sec. \ref{sec:loss}).

\vspace{-0.2cm}\paragraph{Learning prior over a-modal HOI geometry.} Diffusion models are a class of probabilistic generative models that gradually transform a noise from a tractable distribution (Gaussian) to a complex (e.g. real image) data distribution.   Diffusion models are supervised to capture the likelihood by de-noising corrupted images.  During training, they take in corrupted images with a certain amount of noise $\sigma_i$ along with conditions and learn to reconstruct the signals~\cite{ho2020denoising}: 
\begin{align}
    \mathcal{L}_{\text{DDPM}}[\bm x; \bm c] 
    = \mathbb E_{\epsilon\sim \mathcal{N}(\bm 0, \bm I), i} \|\bm x - D_{\bm\psi}(\bm x_{i}, \sigma_i, \bm c) \|_2^2
\end{align}
where $\bm x_i$ is a linear combination of signal $\bm x$ and noise $\epsilon$ while $D_{\bm\psi}$ is the denoiser. 

In our case, as shown in Fig.~\ref{fig:diffusion},  the diffusion model denoises the a-modal geometry rendering of an object given text prompt and hand. Additionally, the diffusion model is also conditioned on the rendering of uv-coordinate of MANO hand $U_h$ because it can better disambiguate if the hand palm faces front or back. More specifically, the training objective is $\mathcal L_{\text{diff}} = \mathcal{L}_{\text{DDPM}}[G_o; C, G_h, U_h]$. The text prompt comes from a text template: ``an image of a hand holding \textit{\{category\}}".

\vspace{-0.3cm}\paragraph{Implementation Details.} When we train the diffusion model with the rendering of ground truth HOI, we draw viewpoints with rotation from the uniform distribution in  $\text{SO}(3)$ .  We use the backbone of  a text-to-image model~\cite{nichol2021glide} with cross attention and modify it to diffuse 5-channel geometry images (3 for normal, 1 for mask and 1 for depth). We initialize the weights from the image-conditioned diffusion model~\cite{nichol2021glide} pretrained with large-scale text-image pairs. The additional channels in the first layer are loaded from the average of the pretrained weights. 

\begin{figure*}
    \centering
\includegraphics[width=\linewidth]{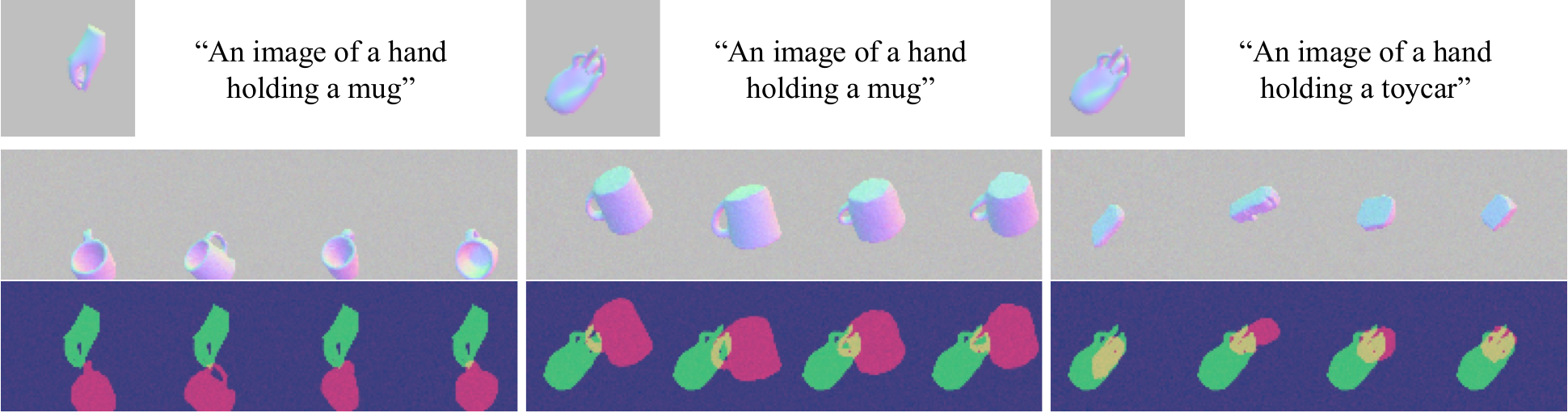}    
    \caption{\textbf{Generations from conditional diffusion model:} Given the geometry rendering of hand $G_h$ (only showing surface normals) and a text prompt $C$,  we visualize 4 different generations from the diffusion model. Middle row shows the generated surface normal of the objects and bottom row visualizes the generated object masks overlayed on the given hand masks. Note the left and middle column share the same text condition while middle and right column share the same hand condition.}
    \label{fig:gen}
\end{figure*}

\subsection{Reconstructing Interaction Clips in 3D}
\label{sec:loss}
After learning the above interactions prior, at inference time when given a short monocular clip with semantic masks of hand and object, we optimize a per-sequence HOI representation to recover the underlying hand-object interactions. We do so by differentiable rendering of the 3D scene representation from the original views and from random novel views. The optimization objectives consist of the following terms. 

\vspace{-0.3cm}\paragraph{Reprojection error.} First, the HOI representation is optimized to explain the input video. We render the semantic mask of the scene from the estimated cameras for each frame and compare the rendering of the semantic masks (considering hand-object occlusion ) with the ground truth masks:
$\mathcal{L}_{\text{reproj}} = \sum_t \| M^t - \hat M^t \|_1$

\vspace{-0.3cm}\paragraph{Learned prior guidance.} 
In the meantime,  the scene is guided by the learned interactino prior to appear more likely from a novel viewpoint following Scored Distillation Sampling (SDS) \cite{poole2022dreamfusion}. SDS treats the output of a diffusion model  as a critic to approximate the gradient step towards more likely images without back-propagating through the  diffusion model for compute efficiency: 
\begin{align}
    \mathcal{L}_{SDS} = \mathbb{E}_{v,\epsilon, i} [w_i\|\pi(\bm\phi^t) - \hat G_o^i\|_2^2]
\end{align}
where $\hat G^i_o$ is the reconstructed signal from the pre-trained diffusion model. Please refer to relevant works~\cite{melaskyriazi2023realfusion,poole2022dreamfusion} or supplmentary for full details.

\vspace{-0.3cm}\paragraph{Other regularization.} We also include two regularization terms: one Eikonal loss~\cite{icml2020_2086} that encourages the implicit field $\bm\phi$ to be a valid distance function $\mathcal L_{\text{eik}} = \|\nabla_X \bm\phi^2 - 1 \|^2$, and another temporal loss that encourages the hand to  move smoothly with respect to the object $\mathcal L_{\text{smooth}} = \sum_t \|T^t_{h\to o} H^t - T^{t-1}_{h\to o}H^{t-1}\|_2^2 $

\vspace{-0.3cm}\paragraph{Initialization and training details.} 
While the camera and object poses are learned jointly with object shape, it is crucial to initialize them to a coarse position \cite{lin2021barf}.  We use FrankMocap~\cite{rong2021frankmocap}, an off-the-shelf hand reconstruction system, to initialize the hand parameters, camera-to-hand transformations, and camera intrinsic. More specifically, FrankMocap predicts finger articulation $\bm\theta_A^t$, wrist orientation $\bm\theta_w^t$, and a weak perspective camera.  The last two are used to compute camera-to-hand transformation and intrinsics of a full perspective camera. See appendix for derivation.  We initialize the object implicit field to a coarse sphere~\cite{yariv2021volume} and the object poses $T^t_{h\to o}$ to identity such that the initial object is roughly round hand palm.

The per-frame hand pose estimation sometimes fails miserably in some challenging frames due to occlusion and motion blur. We run a lightweight trajectory optimization on wrist orientation to correct the catastrophic failure. The optimization objective encourages smooth joint motion across frames while penalizing the difference to the per-frame prediction, \ie $\mathcal L  = \|H(\bm x^t) - H(\hat{\bm x}^t)\| + \lambda \| H(\bm x^{t+1}) - H(\bm x^t) \|$ where $\lambda$ is $0.01$. Please see appendix for full details.

\section{Experiment}


\definecolor{first}{rgb}{1.0, .83, 0.3}
\definecolor{second}{rgb}{1.0, 0.93, 0.7}
\def \first {\cellcolor{first}}
\def \second {\cellcolor{second}}
\def \third {}

\begin{table*}[t]
\footnotesize
\begin{center}
\caption{\textbf{Comparison with baselines:}  We compare our method along with prior works HHOR \cite{huang2022hhor} and iHOI \cite{ye2022hand} on the HOI4D dataset and report object reconstruction error in $F@5$mm and $F@10$mm scores and Chamfer Distance ($CD$). }

\label{tab:hoi4d}

\setlength{\tabcolsep}{2pt}
\resizebox{\linewidth}{!}{
\begin{tabular}{l ccc ccc ccc ccc ccc ccc ccc}
\toprule

& \multicolumn{3}{c}{Mug} 
& \multicolumn{3}{c}{Bottle}
& \multicolumn{3}{c}{Kettle} 
& \multicolumn{3}{c}{Bowl} 
& \multicolumn{3}{c}{Knife} 
& \multicolumn{3}{c}{ToyCar} 
& \multicolumn{3}{c}{Mean}\\

\cmidrule(r){2-4} \cmidrule(r){5-7} \cmidrule(r){8-10} \cmidrule(r){11-13} \cmidrule(r){14-16}
\cmidrule(r){17-19} \cmidrule(r){20-22}

& $F@5$  & $F@10$ & $CD$  
& $F@5$  & $F@10$ & $CD$  
& $F@5$  & $F@10$ & $CD$  
& $F@5$  & $F@10$ & $CD$  
& $F@5$  & $F@10$ & $CD$  
& $F@5$  & $F@10$ & $CD$  
& $F@5$  & $F@10$ & $CD$  
\\

\midrule

HHOR~\cite{huang2022hhor} & 0.18 & 0.37 & 7.0 & 0.26 & 0.56 & 3.1 & 0.12 & 0.30 & 11.3 & 0.31 & 0.54 & 4.2 & \first \textbf{0.71} & \second 0.93 & \first \textbf{0.6} & 0.26 & 0.59 & 1.9 & 0.31 & 0.55 & 4.7 \\
iHOI~\cite{ye2022hand} & \second 0.44 & \second 0.71 & \second 2.1 & \second 0.48 & \second 0.77 & \second 1.5 & \second 0.21 & \second 0.45 & \second 6.3 & \second 0.38 & \second 0.64 & \second 3.1 & 0.33 & 0.68 & 2.8 & \second 0.66 & \second 0.95 & \second 0.5 & \second 0.42 & \second 0.70 & \second 2.7 \\
Ours (\ours)  & \first \textbf{0.64} & \first \textbf{0.86} & \first \textbf{1.0} & \first \textbf{0.54} & \first \textbf{0.92} & \first \textbf{0.7} & \first \textbf{0.43} & \first \textbf{0.77} & \first \textbf{1.5} & \first \textbf{0.79} & \first \textbf{0.98} & \first \textbf{0.4} & \second 0.50 & \first \textbf{0.95} & \second 0.8 & \first \textbf{0.83} & \first \textbf{0.99} & \first \textbf{0.3} & \first \textbf{0.62} & \first \textbf{0.91} & \first \textbf{0.8} \\




\bottomrule
\end{tabular}
}

\end{center}

\end{table*}

\begin{figure*}
    \centering
    \includegraphics[width=\linewidth]{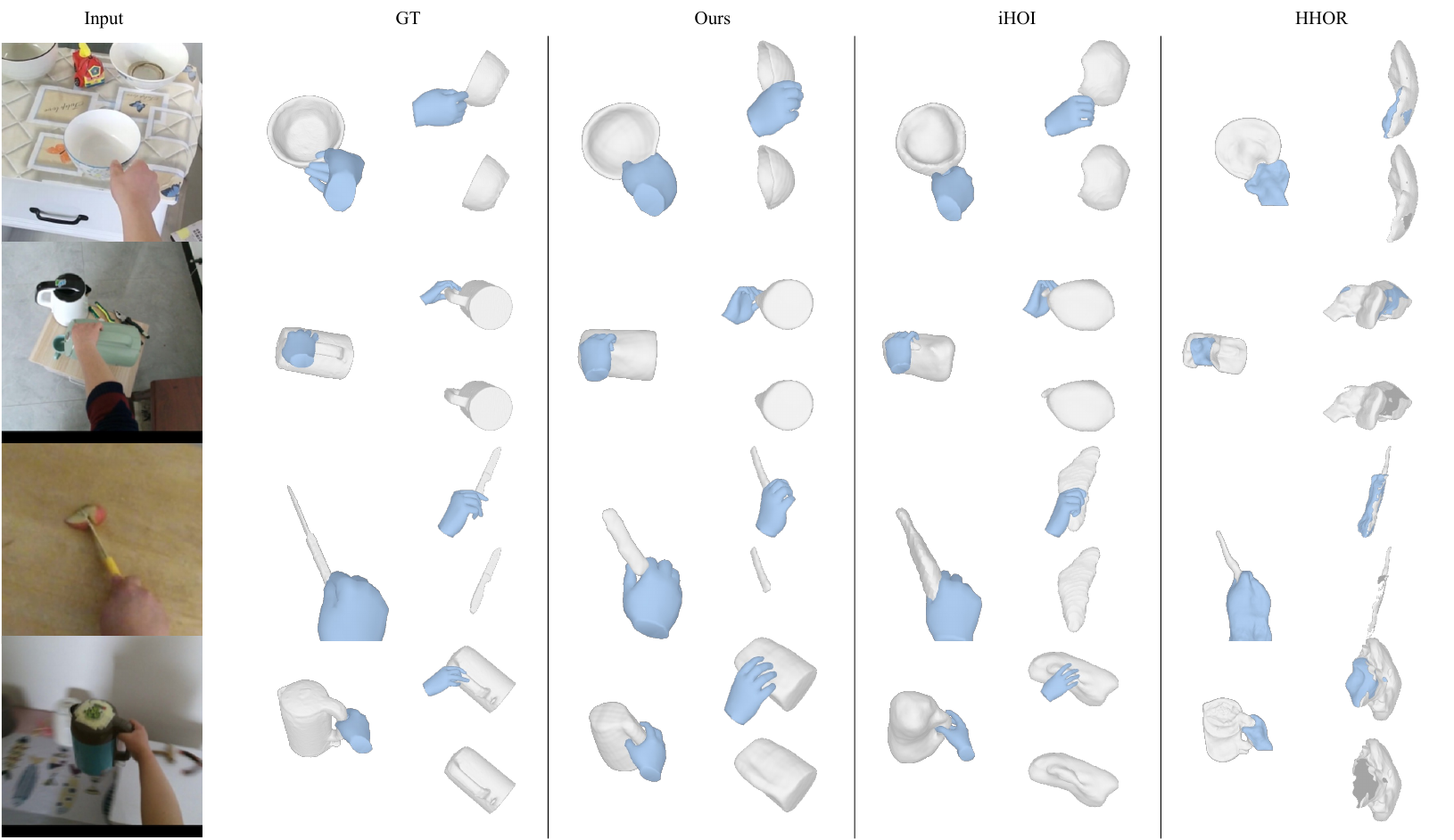}
    \caption{\textbf{Qualitative evaluation on HOI4D: } We show reconstruction by our method (\ours) along with two baselines \cite{huang2022hhor, ye2022hand} in the image frame (left) and another novel view with (top right) or without (bottom right) hand. Please see project website for reconstruction videos. }
    \label{fig:hoi4d}
\end{figure*}

We first train the diffusion model on the egocentric HOI4D~\cite{hoi4d} dataset and visualize its generations  in Section~\ref{sec:vis_prior}. Then, 
we evaluate the reconstruction of hand-object interactions quantitatively and qualitatively on the held-out sequences 
and compare \ours~ with two model-free baselines (Section~\ref{sec:hoi4d_result}).  We then analyze the effects of both category-prior and hand-prior respectively, ablate the contribution from each geometry modality, and analyze its robustness to initial prediction errors (Section~\ref{sec:ablation}). In Section~\ref{sec:template}, we discuss how \ours~ compares with other template-based methods. Lastly, in Section~\ref{sec:wild}, we show that our method is able to reconstruct HOI from in-the-wild video clips both in first-person and from third-person view. 

\vspace{-0.5cm}\paragraph{Dataset and Setup.}
 HOI4D is an egocentric dataset consisting of  short video clips of hand interacting with objects. It is collected under controlled environments and recorded by head-wear RGBD cameras. Ground truth is provided by fitting 6D pose of scanned objects to the RGBD videos.  We use all of  the 6 rigid object categories in portable size (mug, bottle, kettle, knife, toy car, bowl). To train the diffusion model, we render one random novel viewpoint for each frame resulting in 35k training points. 
We test the object reconstruction on held-out instances, two sequences per category. All of baselines and our method use the segmentation masks from ground truth annotations and the hand poses from the off-the-shelf prediction system~\cite{rong2021frankmocap} if required.

For in-the-wild dataset, we test on clips from EPIC-KITCHENS~\cite{Damen2018EPICKITCHENS} videos and casual YouTube videos downloaded from the Internet. The segmentation masks are obtained using an off-the-shelf video object segmentation system~\cite{cheng2021stcn}.

\subsection{Visualizing Data-Driven Priors}
\label{sec:vis_prior}
We show conditional generations by the pre-trained diffusion model in Fig.~\ref{fig:generation}.  Given the geometry rendering of hand (only visualizing surface normal), as well as a text prompt, we visualize 4 different generations from the diffusion model. Middle row shows the generated surface normal of the object and bottom row visualizes the generated object masks overlayed on top of the given hand mask, for a better view of the hand-object relations. Note that left and middle column condition on the same text prompts while middle and right column conditions on the same hand pose. Please see appendix for additional examples and visualizations of all modalities. 

The generated object match the category information in the prompt while the generations are diverse in position, orientation, and size. 
Yet, all of the hand-object interactions are plausible, \eg different generated handles all appear at the tip of the hand. 
Comparing middle and right examples, different category prompts lead to different generations given the same hand rendering. 
With the same prompt but different hands (left and middle), the generated objects flip the orientation accordingly. 
In summary, Fig.~\ref{fig:generation} indicates that the learned prior is aware of both the hand  prior and the category-level prior hence being informative to guide the 3D reconstruction from clips.

\subsection{Comparing Reconstructions of HOI4D}
\label{sec:hoi4d_result}
\paragraph{Evaluation Metric.} We evaluate the object reconstruction errors. Following prior works~\cite{huang2022hhor,Hampali2022InHand3O}, we first align the reconstructed object shape with the ground truth by Iterative Closest Point (ICP), allowing scaling. Then we compute Chamfer distance (CD), F-score~\cite{tatarchenko2019single} at $5mm$ and $10mm$ and report mean over 2 sequences for each category. Chamfer distance focuses on the global shapes more and is affected by outliers while F-score focuses on local shape details at different thresholds~\cite{tatarchenko2019single}. 

\vspace{-0.5cm}\paragraph{Baselines.}
While few prior works tackle our challenging setting -- 3D HOI reconstruction from casual monocular clips without knowing the templates, the closest works are two template-free methods from Huang  \etal \cite{huang2022hhor} (HHOR) and Ye \etal \cite{ye2022hand} (iHOI). 

HHOR is proposed for in-hand scanning. It  optimizes a  deformable semantic implicit field to jointly model hand and object. HHOR captures the dynamics by a per-frame warping field while no prior is used during optimization. 
iHOI is a feed-forward method and reconstructs 3D  objects from single-view images by learning the hand prior between hand poses and object shapes. The method does not leverage category-level prior and do not consider time-consistency of shapes. 
We finetune their pretrained model to take in segmentation masks. We  evaluate their result by aligning their predictions with ground truth for each frame and report the average number across all frames.


\definecolor{first}{rgb}{1.0, .83, 0.3}
\definecolor{second}{rgb}{1.0, 0.93, 0.7}
\def \first {\cellcolor{first}}
\def \second {\cellcolor{second}}
\def \third {}

\begin{table}[t!]
\footnotesize
\begin{center}
\caption{\textbf{Analysis of the effect of data-driven priors: } 
Quantitative results on HOI4D for object reconstruction error in the object-centric frame ($F@5$, $F@10$,
$CD$) and for hand-object alignment in the hand frame ($CD_h$). We compare our method  with ablations that does not use prior, or use other variants of diffusion models that only conditions on hand or category. \vspace{.5em}}
\label{tab:prior}
\setlength{\tabcolsep}{2pt}
\resizebox{0.65\linewidth}{!}{

\begin{tabular}{l c c c | c }
\toprule
& $F@5$ & $F@10$ & $CD$ & $CD_{h}$ \\




\midrule
No prior & 0.47 & 0.73 & 2.7 & \first \textbf{37.0} \\
Hand prior & 0.39 & 0.65 & 2.8 & 55.0 \\
Category prior & \second 0.56 & \second 0.87 & \second 1.6 & 85.2 \\
Ours  & \first \textbf{0.62} & \first \textbf{0.91} & \first \textbf{0.8} & \second 48.7 \\


\bottomrule
\vspace{-3.5em}
\end{tabular}
}

\end{center}

\end{table}

\begin{figure}
    \centering
    \includegraphics[width=\linewidth]{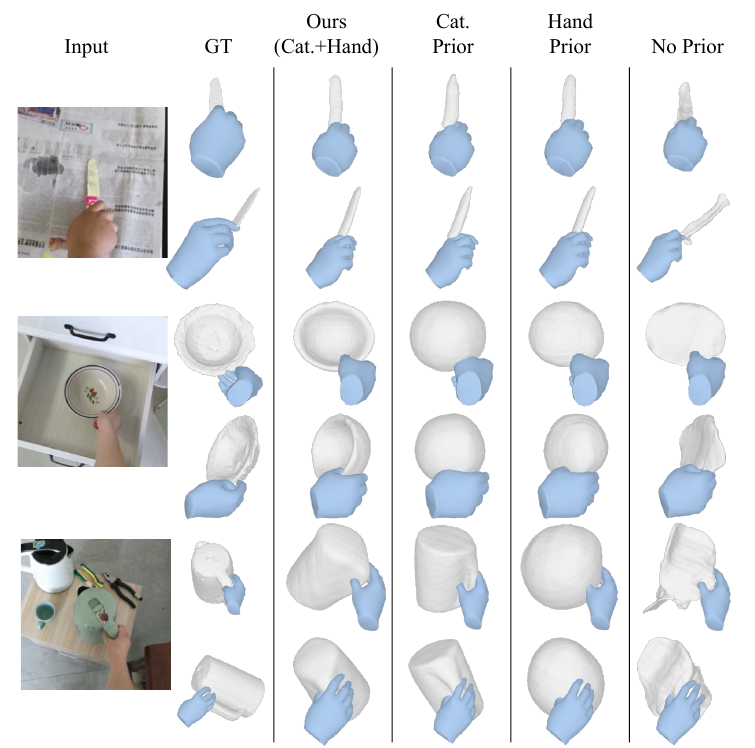}
    \caption{ \textbf{Ablation Study:}
    We show reconstruction in the image frame (top) and from a novel view (bottom) by our method along with 
    ablations using other variants of diffusion models that only conditions on category or hand, and one that does not use prior.
    }
    \label{fig:prior}
\end{figure}


\definecolor{first}{rgb}{1.0, .83, 0.3}
\definecolor{second}{rgb}{1.0, 0.93, 0.7}
\def \first {\cellcolor{first}}
\def \second {\cellcolor{second}}
\def \third {}

\begin{table}[t!]
\footnotesize
\begin{center}
\caption{\textbf{Ablation without surface normal, mask and depth in distillation:}
Quantitative results on HOI4D for object reconstruction error in the object-centric frame ($F@5$, $F@10$,
$CD$) and for hand-object alignment in the hand frame ($CD_h$). We compare our method with other ablations that do not distill normals, masks, and depths respectively. \vspace{.5em}}
\label{tab:weight}
\setlength{\tabcolsep}{2pt}
\resizebox{0.6\linewidth}{!}{

\begin{tabular}{l c c c | c }
\toprule
& $F@5$ & $F@10$ & $CD$ & $CD_{h}$ \\




\midrule
$-$ normal & 0.36 & 0.58 & 4.3 & 220.2 \\
$-$ mask & 0.56 & 0.82 & 1.3 & 128.1 \\
$-$ depth & \first \textbf{0.66} & \second 0.90 & \second 0.9 & \second 88.0 \\
Ours & \second 0.62 & \first \textbf{0.91} & \first \textbf{0.8} & \first \textbf{48.7} \\

\bottomrule
\vspace{-2.5em}
\end{tabular}
}

\end{center}

\end{table}

\begin{figure}
    \centering
    \includegraphics[width=\linewidth]{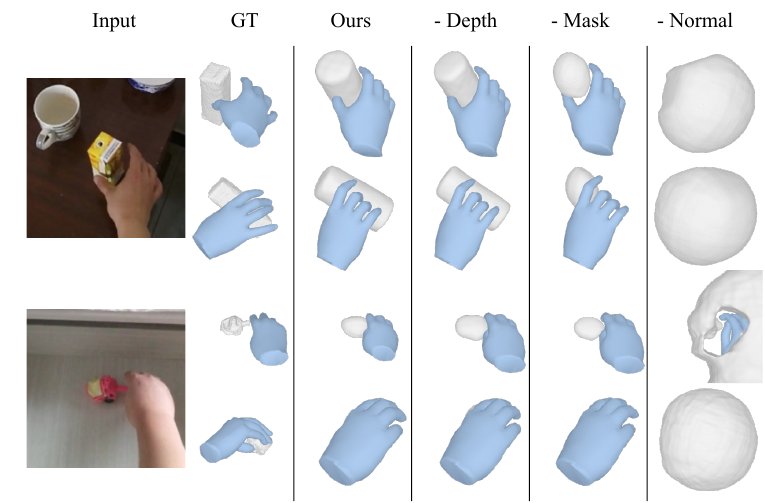}
    \caption{\textbf{Ablation Study:} 
    We show reconstruction in the image frame (top) and from a novel view (bottom) by our method along with 
    other variants that do not distill on depth, mask, and normals.}
    \label{fig:weight}
\end{figure}

\vspace{-0.5cm}\paragraph{Results.} We visualize the reconstructed HOI and  object shapes from the image frame and a novel viewpoint in Fig.~\ref{fig:hoi4d}. HHOR generates good-looking results from the original view but actually degenerates to a flat surface since it does not incorporate any prior knowledge besides the visual observation. It also cannot decompose the hand and the object on the unobserved side of the scene because HHOR distinguishes them by per-point classification predicted from the neural field, which does not get gradient from the observations.  iHOI reconstructs better object shapes and interactions but it is not very accurate as it cannot aggregate information across different frames. Its prediction is not time consistent either (better visualized as videos). In contrast, we are able to reconstruct time-persistent object shapes with time changing hand poses. The reconstructed object is more accurate, \eg knife blade is thinner and the kettle body is more cylindrical. 

This is consistent with quantitative results in Tab.~\ref{tab:hoi4d}.  HHOR generally performs unfavorably except for knife category. While iHOI performs better, its quality is limited by only relying on information from a single frame. \ours~ outperforms the baseline methods by large margins in most sequences and performs the best on all three metrics for mean values.

\subsection{Ablation Studies}
\label{sec:ablation}
We ablate our system carefully to analyze the contribution of each component. 
Besides the object reconstruction errors  in the aligned object-centric frame, we further evaluate the hand-object \textit{arrangement} by reporting the Chamfer distance of objects in hand frame, \ie $CD_h \equiv CD(T^t_{o\to h} O, \hat T^t_{o\to h} \hat O)$). We only report mean value in the main paper. Please refer to supplementary for category-wise results. 

\vspace{-0.5cm}\paragraph{How does each learned prior help? } We analyze how the category and hand priors affect reconstruction by training  two more diffusion models conditioned only on text-prompt or hand renderings respectively. We also compare with the variant without optimizing $\mathcal L_{\text{SDS}}$ (no prior). As reported quantitatively, we find that  \textit{category prior helps object reconstructions  while hand prior helps hand-object relation (Tab.~\ref{tab:prior}). } And combining them both results in best performance. 


We highlight an interesting qualitative result of reconstructing the bowl in  Fig.~\ref{fig:prior}. Neither prior can reconstruct the concave shape on its own -- the hand pose alone is not predictive enough of the object shape while only knowing the object to be a bowl cannot make the SDS converge to  a consensus direction that the bowl faces. Only knowing \textit{both} can the concave shapes be recovered. This example further highlights the importance of both priors.


\vspace{0.3cm}\noindent\textbf{Which geometry modality matters more for distillation? }
Next, we investigate how much each geometry modality (mask, normal, depth) contributes when distilling them into 3D shapes. Given the same pretrained diffusion model, we disable one of the three input modalities in optimization by setting its weight on $\mathcal L_{\text{SDS}}$ to 0. 

As visualized in Fig.~\ref{fig:weight}, the surface normal is the most important modality. Interestingly, the model collapses if not distilling surface normals and even performs worse than the no-prior variant. Without distillation on masks, the object shape becomes less accurate probably because binary masks predict more discriminative signals on shapes. Relative depth does not help much with global object shape but it helps in aligning detailed local geometry ($F@5$) and aligning the object to hand ($F@10$).

\vspace{-0.5cm}\paragraph{How robust is the system to hand pose prediction errors?}
\begin{table}[t!]
\centering
\caption{\textbf{Error analysis against hand pose noise:} * marks our unablated method. Numbers in parentheses are per-frame prediction errors before optimization.\vspace{.5em} }
\label{tab:robust}
\footnotesize
\setlength{\tabcolsep}{2pt}
\resizebox{0.9\linewidth}{!}{    
    \begin{tabular}{l ccc cc}
    \toprule
    & \multicolumn{3}{c}{Object Reconstruction} & \multicolumn{2}{c}{Hand Estimation} \\
    \cmidrule(r){2-4}  \cmidrule(r){5-6} 
& $F@5$ $\uparrow$ & $F@10$ $\uparrow$ & $CD$ $\downarrow$   &   MPJPE $\downarrow$  & AUC $\uparrow$\\
\midrule
GT & 0.68 & 0.91 & 0.75 &	 -- & --\\
Prediction* & 0.62 &	0.91	& 0.77  &   26.9(28.4)  &  0.49(0.47) \\
Pred. Error $\times 2$ & 0.63 & 0.87 & 1.01 &  40.7(44.6) & 0.31(0.27) \\
\bottomrule
\vspace{-2.5em}
\end{tabular} 
}
\end{table}

We report the object reconstruction performance when using GT vs predicted hand pose in Tab.~\ref{tab:robust}, and find that our system is robust to some prediction error. Moreover, even if we artificially degrade the prediction by doubling the error, our performance remains better than the baselines (Tab.~\ref{tab:hoi4d}). We also report the hand pose estimation metrics and find that our optimization improves the initial predictions (in parentheses).

\subsection{Comparing with Template-Based Methods}
\label{sec:template}

\begin{table}[t!]
\centering
\caption{\textbf{Comparison with template-based baseline:} Quantitative results on the HOI4D dataset for object
reconstruction error in the object-centric frame ($F@5$, $F@10$, $CD$) and for hand-object alignment ($CD_h$). We compare our method with HOMAN~\cite{hasson2021towards} with the ground truth template (-GT), with random templates from the training split (and reporting the average), and with furthest template from the ground truth (-furthest). \vspace{.5em}}
\label{tab:template}
\footnotesize
\setlength{\tabcolsep}{2pt}
\resizebox{0.8\linewidth}{!}{    
    \begin{tabular}{l ccc |c}
    \toprule
& $F@5$ $\uparrow$ & $F@10$ $\uparrow$ & $CD$ $\downarrow$ & $CD_h$ $\downarrow$   \\
\midrule
HOMAN-GT & 1.00 & 1.00 & 0.00 & 84.3 \\
HOMAN-average & 0.76	& 0.94	& 0.48 & 120.9 \\ 
HOMAN-furthest & 0.49 &	0.78  &	1.33 &  157.9  \\
Ours(\ours) & 0.62 &	0.91	& 0.78 & 48.7 \\
\bottomrule
\end{tabular} 
}
\vspace{2mm}
\vspace{-5mm}
\end{table}
\begin{figure}
    \centering
    \includegraphics[width=\linewidth]{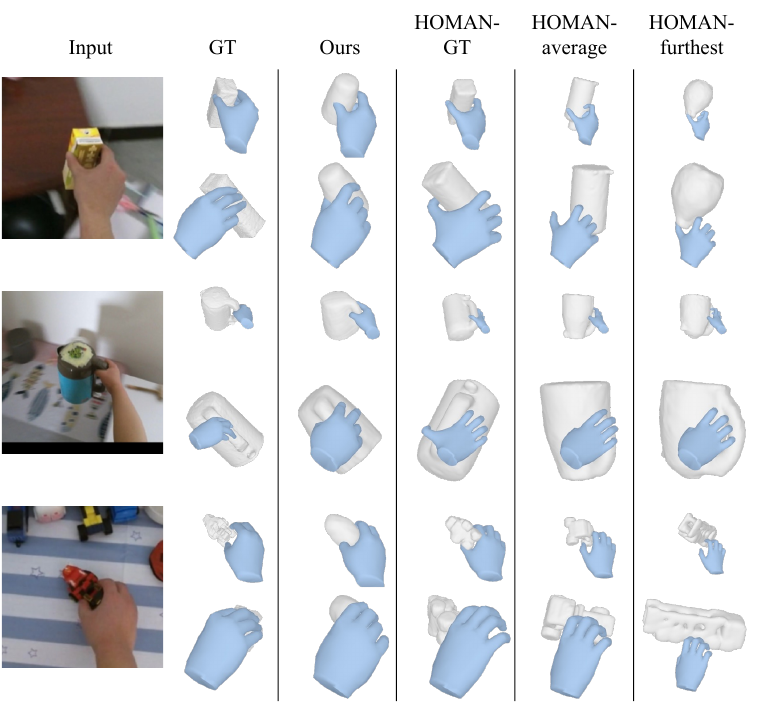}
    \caption{\textbf{Comparing with template-based method:}
    We show reconstruction in the image frame (top) and from a novel view (bottom) by our method and
    HOMAN~\cite{hasson2021towards} when provided with ground-truth templates, a random template, and the most dissimilar template in the training split.}
    \label{fig:template}
\end{figure}

\begin{figure*}
    \centering
    \includegraphics[width=\linewidth]{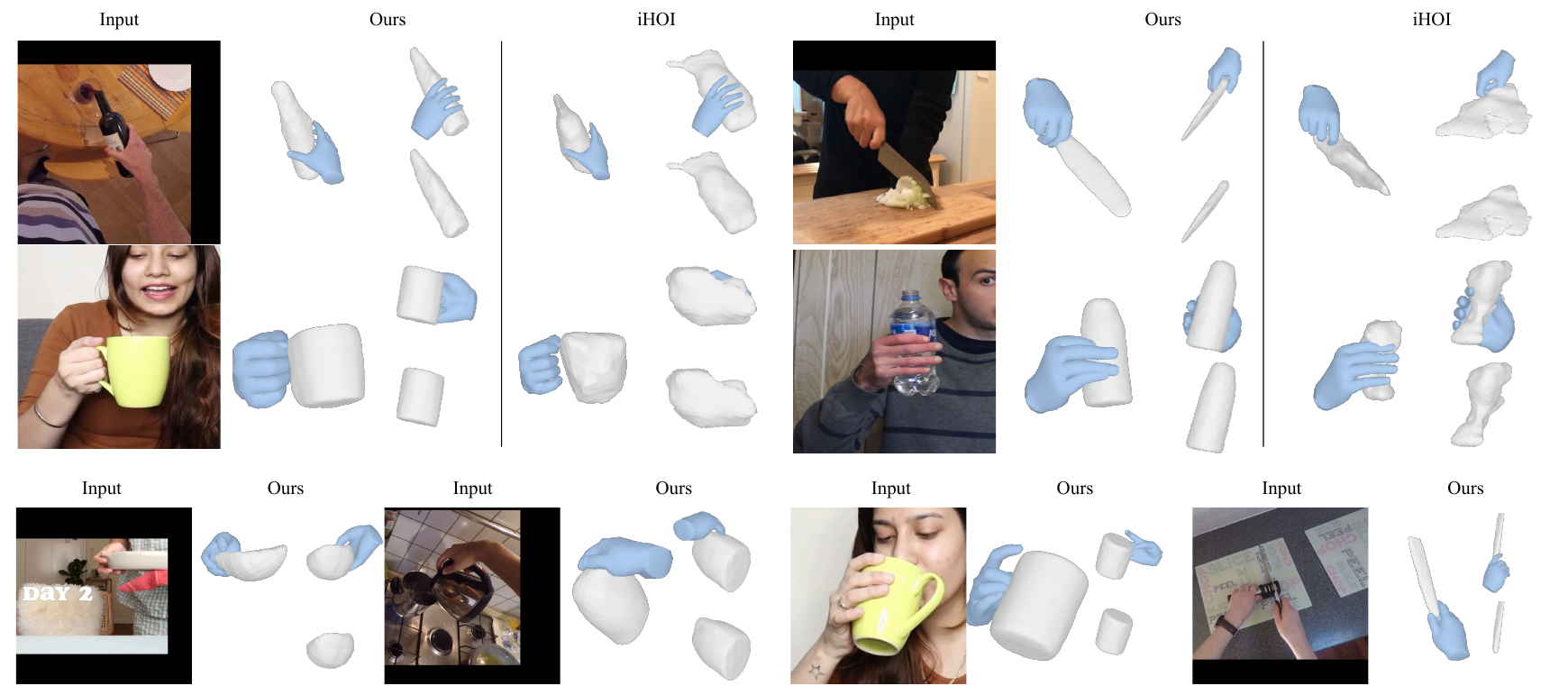}
    \caption{\textbf{Qualitative evaluation on in-the-wild video clips:} We show reconstruction by our method (\ours) and iHOI~\cite{ye2022hand} in the image frame (left) and a novel view with (top right) or without (bottom right) hand. Please see project
website for reconstruction videos.}
    \label{fig:wild}
\end{figure*}

We compare with HOMAN~\cite{hasson2021towards}, a representative template-based method that optimizes object 6D poses and hand articulations with respect to reprojection error and multiple interaction objectives including contact, intersection, distance, relative depth,  temporal smoothness, \etc.

We show quantitative and qualitative results in Tab.~\ref{tab:template} and ~\ref{fig:template}. Note that evaluating HOMAN in terms of object reconstruction is equivalent to evaluating templates since the objects are aligned in the object-centric frame. We first report the average  object reconstruction errors when optimizing with different templates from training sets.  While the gap indicates potential room to improve object shapes for template-free methods, \ours~is favorable over some templates in the training set. Nevertheless, when evaluating the objects in the hand frame, \ours~outperforms HOMAN by a large margin. 
The numbers along with visualizations in Fig.~\ref{fig:template} indicate that template-based methods, even when optimizes with multiple objectives to encourage interactions, still struggle to place objects in the context of hands, especially for subtle parts like handles.  Furthermore, optimizing with random templates degrades $CD_h$ significantly, highlighting the inherent drawbacks of template-based methods to demand the accurate templates.


\subsection{Reconstructing In-the-Wild Video Clips}
\label{sec:wild}

Lastly, we show that our method can be directly applied to more challenging video clips.  
In Fig.~\ref{fig:wild} top, we compare between our method and iHOI~\cite{ye2022hand}. iHOI predicts reasonable shapes from the front view but  fails on transparent objects like the plastic bottle since it is never trained on such appearance. In contrast, we transfer better to in-the-wild sequences as the learned prior only take on geometry cues. In Fig.~\ref{fig:wild} bottom, we visualize more results from our method. By incorporating learned priors, our method is robust to mask prediction inaccuracy, occlusion from irrelevant objects (the onion occludes knife blade), truncation of the HOI scene (bowl at the bottom left), \etc. 
Our method can also work across ego-centric and third-person views since the learned prior is trained with uniformly sampled viewpoints. The reconstructed shapes vary from thin objects like knives to larger objects like kettles.

\section{Conclusion}
In this work, we propose a method to reconstruct hand-object interactions without any object templates from daily video clips. Our method is the first to tackle this challenging setting.  We represent the HOI scene by a model-free implicit field for the object and a model-based mesh for the hand. The scene is  optimized with respect to re-projection error and a data-driven geometry prior that captures the object shape given category information and hand poses. Both of these modules are shown as critical for successful reconstruction.  Despite the encouraging results, there are several limitations: the current method can only handle small hand-object motions in short video clips up to a few ($\sim$5) seconds; the reconstructed objects still miss details of shape.  Despite the challenges, we believe that our work takes an encouraging step towards a holistic understanding of human-object interactions in everyday videos. 

\paragraph{Acknowledgements.}
The authors would thank Di Huang for HHOR comparison. We thank Dandan Shan, Sudeep Dasari for helping with EPIC-KITCHENS datasets. We also thank Sudeep Dasari, Hanzhe Hu, Helen Jiang for detailed feedback on the manuscript. Yufei was partially supported by the NVIDIA fellowship.
{\small
\bibliographystyle{ieee_fullname}
\bibliography{ref}

\begin{thebibliography}{10}\itemsep=-1pt

\bibitem{bhatnagar2022behave}
Bharat~Lal Bhatnagar, Xianghui Xie, Ilya~A Petrov, Cristian Sminchisescu,
  Christian Theobalt, and Gerard Pons-Moll.
\newblock Behave: Dataset and method for tracking human object interactions.
\newblock In {\em CVPR}, 2022.

\bibitem{brahmbhatt2020contactpose}
Samarth Brahmbhatt, Chengcheng Tang, Christopher~D Twigg, Charles~C Kemp, and
  James Hays.
\newblock Contactpose: A dataset of grasps with object contact and hand pose.
\newblock In {\em ECCV}, 2020.

\bibitem{brooks2022hallucinating}
Tim Brooks and Alexei~A Efros.
\newblock Hallucinating pose-compatible scenes.
\newblock In {\em ECCV}, 2022.

\bibitem{Cao2020ReconstructingHI}
Zhe Cao, Ilija Radosavovic, Angjoo Kanazawa, and Jitendra Malik.
\newblock Reconstructing hand-object interactions in the wild.
\newblock {\em ICCV}, 2021.

\bibitem{rhoi2020}
Zhe Cao, Ilija Radosavovic, Angjoo Kanazawa, and Jitendra Malik.
\newblock Reconstructing hand-object interactions in the wild.
\newblock {\em ICCV}, 2021.

\bibitem{cao2021reconstructing}
Zhe Cao, Ilija Radosavovic, Angjoo Kanazawa, and Jitendra Malik.
\newblock Reconstructing hand-object interactions in the wild.
\newblock In {\em ICCV}, 2021.

\bibitem{chen2023gsdf}
Zerui Chen, Shizhe Chen, Cordelia Schmid, and Ivan Laptev.
\newblock gsdf: Geometry-driven signed distance functions for 3d hand-object
  reconstruction.
\newblock In {\em CVPR}, 2023.

\bibitem{chen2022alignsdf}
Zerui Chen, Yana Hasson, Cordelia Schmid, and Ivan Laptev.
\newblock Alignsdf: Pose-aligned signed distance fields for hand-object
  reconstruction.
\newblock In {\em ECCV}, 2022.

\bibitem{cheng2021stcn}
Ho~Kei Cheng, Yu-Wing Tai, and Chi-Keung Tang.
\newblock Rethinking space-time networks with improved memory coverage for
  efficient video object segmentation.
\newblock In {\em NeurIPS}, 2021.

\bibitem{chuang2018learning}
Ching-Yao Chuang, Jiaman Li, Antonio Torralba, and Sanja Fidler.
\newblock Learning to act properly: Predicting and explaining affordances from
  images.
\newblock In {\em CVPR}, 2018.

\bibitem{corona2020ganhand}
Enric Corona, Albert Pumarola, Guillem Alenya, Francesc Moreno-Noguer, and
  Gr{\'e}gory Rogez.
\newblock Ganhand: Predicting human grasp affordances in multi-object scenes.
\newblock In {\em CVPR}, 2020.

\bibitem{Damen2018EPICKITCHENS}
Dima Damen, Hazel Doughty, Giovanni~Maria Farinella, Sanja Fidler, Antonino
  Furnari, Evangelos Kazakos, Davide Moltisanti, Jonathan Munro, Toby Perrett,
  Will Price, and Michael Wray.
\newblock Scaling egocentric vision: The epic-kitchens dataset.
\newblock In {\em ECCV}, 2018.

\bibitem{Deng2022NeRDiSN}
Congyue Deng, Chiyu~Max Jiang, C. Qi, Xinchen Yan, Yin Zhou, Leonidas~J.
  Guibas, and Drago Anguelov.
\newblock Nerdi: Single-view nerf synthesis with language-guided diffusion as
  general image priors.
\newblock {\em CVPR}, 2023.

\bibitem{fan2023arctic}
Zicong Fan, Omid Taheri, Dimitrios Tzionas, Muhammed Kocabas, Manuel Kaufmann,
  Michael~J. Black, and Otmar Hilliges.
\newblock {ARCTIC}: A dataset for dexterous bimanual hand-object manipulation.
\newblock In {\em CVPR}, 2023.

\bibitem{gao2022monocular}
Hang Gao, Ruilong Li, Shubham Tulsiani, Bryan Russell, and Angjoo Kanazawa.
\newblock Monocular dynamic view synthesis: A reality check.
\newblock {\em NeurIPS}, 2022.

\bibitem{garcia2018first}
Guillermo Garcia-Hernando, Shanxin Yuan, Seungryul Baek, and Tae-Kyun Kim.
\newblock First-person hand action benchmark with rgb-d videos and 3d hand pose
  annotations.
\newblock In {\em CVPR}, 2018.

\bibitem{icml2020_2086}
Amos Gropp, Lior Yariv, Niv Haim, Matan Atzmon, and Yaron Lipman.
\newblock Implicit geometric regularization for learning shapes.
\newblock In {\em PMLR}, 2020.

\bibitem{guo2019relightables}
Kaiwen Guo, Peter Lincoln, Philip Davidson, Jay Busch, Xueming Yu, Matt Whalen,
  Geoff Harvey, Sergio Orts-Escolano, Rohit Pandey, Jason Dourgarian, et~al.
\newblock The relightables: Volumetric performance capture of humans with
  realistic relighting.
\newblock {\em ToG}, (6), 2019.

\bibitem{hampali2023hand}
Shreyas Hampali, Tomas Hodan, Luan Tran, Lingni Ma, Cem Keskin, and Vincent
  Lepetit.
\newblock In-hand 3d object scanning from an rgb sequence.
\newblock In {\em CVPR}, 2023.

\bibitem{Hampali2022InHand3O}
Shreyas Hampali, Tom{\'a}s Hodan, Luan Tran, Lingni Ma, Cem Keskin, and Vincent
  Lepetit.
\newblock In-hand 3d object scanning from an rgb sequence.
\newblock {\em CVPR}, 2023.

\bibitem{hampali2020honnotate}
Shreyas Hampali, Mahdi Rad, Markus Oberweger, and Vincent Lepetit.
\newblock Honnotate: A method for 3d annotation of hand and object poses.
\newblock In {\em CVPR}, 2020.

\bibitem{Hasson2020LeveragingPC}
Yana Hasson, Bugra Tekin, Federica Bogo, Ivan Laptev, Marc Pollefeys, and
  Cordelia Schmid.
\newblock Leveraging photometric consistency over time for sparsely supervised
  hand-object reconstruction.
\newblock In {\em CVPR}, 2020.

\bibitem{hasson2021towards}
Yana Hasson, G{\"u}l Varol, Cordelia Schmid, and Ivan Laptev.
\newblock Towards unconstrained joint hand-object reconstruction from rgb
  videos.
\newblock In {\em 3DV}, 2021.

\bibitem{hasson19_obman}
Yana Hasson, G{\"u}l Varol, Dimitris Tzionas, Igor Kalevatykh, Michael~J.
  Black, Ivan Laptev, and Cordelia Schmid.
\newblock Learning joint reconstruction of hands and manipulated objects.
\newblock In {\em CVPR}, 2019.

\bibitem{Ho2020DenoisingDP}
Jonathan Ho, Ajay Jain, and P. Abbeel.
\newblock Denoising diffusion probabilistic models.
\newblock {\em NeurIPS}, 2020.

\bibitem{ho2020denoising}
Jonathan Ho, Ajay Jain, and Pieter Abbeel.
\newblock Denoising diffusion probabilistic models.
\newblock {\em NeurIPS}, 2020.

\bibitem{hu2021lora}
Edward~J Hu, Yelong Shen, Phillip Wallis, Zeyuan Allen-Zhu, Yuanzhi Li, Shean
  Wang, Lu Wang, and Weizhu Chen.
\newblock Lora: Low-rank adaptation of large language models.
\newblock {\em ICLR}, 2021.

\bibitem{hu2022hand}
Hezhen Hu, Weilun Wang, Wengang Zhou, and Houqiang Li.
\newblock Hand-object interaction image generation.
\newblock {\em NeurIPS}, 2022.

\bibitem{huang2022hhor}
Di Huang, Xiaopeng Ji, Xingyi He, Jiaming Sun, Tong He, Qing Shuai, Wanli
  Ouyang, and Xiaowei Zhou.
\newblock Reconstructing hand-held objects from monocular video.
\newblock In {\em SIGGRAPH Asia}, 2022.

\bibitem{jiang2021hand}
Hanwen Jiang, Shaowei Liu, Jiashun Wang, and Xiaolong Wang.
\newblock Hand-object contact consistency reasoning for human grasps
  generation.
\newblock In {\em ICCV}, 2021.

\bibitem{jiang2022neuralhofusion}
Yuheng Jiang, Suyi Jiang, Guoxing Sun, Zhuo Su, Kaiwen Guo, Minye Wu, Jingyi
  Yu, and Lan Xu.
\newblock Neuralhofusion: Neural volumetric rendering under human-object
  interactions.
\newblock In {\em CVPR}, 2022.

\bibitem{jun2023shap}
Heewoo Jun and Alex Nichol.
\newblock Shap-e: Generating conditional 3d implicit functions.
\newblock {\em arXiv}, 2023.

\bibitem{karunratanakul2020grasping}
Korrawe Karunratanakul, Jinlong Yang, Yan Zhang, Michael~J Black, Krikamol
  Muandet, and Siyu Tang.
\newblock Grasping field: Learning implicit representations for human grasps.
\newblock In {\em 3DV}, 2020.

\bibitem{kingma2020method}
Diederik~P Kingma, J~Adam Ba, and J Adam.
\newblock A method for stochastic optimization.
\newblock {\em ICLR}, 2020.

\bibitem{kirillov2020pointrend}
Alexander Kirillov, Yuxin Wu, Kaiming He, and Ross Girshick.
\newblock Pointrend: Image segmentation as rendering.
\newblock In {\em CVPR}, 2020.

\bibitem{kulal2023putting}
Sumith Kulal, Tim Brooks, Alex Aiken, Jiajun Wu, Jimei Yang, Jingwan Lu,
  Alexei~A Efros, and Krishna~Kumar Singh.
\newblock Putting people in their place: Affordance-aware human insertion into
  scenes.
\newblock In {\em CVPR}, 2023.

\bibitem{li2019putting}
Xueting Li, Sifei Liu, Kihwan Kim, Xiaolong Wang, Ming-Hsuan Yang, and Jan
  Kautz.
\newblock Putting humans in a scene: Learning affordance in 3d indoor
  environments.
\newblock In {\em CVPR}, 2019.

\bibitem{li2023dynibar}
Zhengqi Li, Qianqian Wang, Forrester Cole, Richard Tucker, and Noah Snavely.
\newblock Dynibar: Neural dynamic image-based rendering.
\newblock In {\em CVPR}, 2023.

\bibitem{lin2022magic3d}
Chen-Hsuan Lin, Jun Gao, Luming Tang, Towaki Takikawa, Xiaohui Zeng, Xun Huang,
  Karsten Kreis, Sanja Fidler, Ming-Yu Liu, and Tsung-Yi Lin.
\newblock Magic3d: High-resolution text-to-3d content creation.
\newblock {\em CVPR}, 2023.

\bibitem{lin2021barf}
Chen-Hsuan Lin, Wei-Chiu Ma, Antonio Torralba, and Simon Lucey.
\newblock Barf: Bundle-adjusting neural radiance fields.
\newblock In {\em ICCV}, 2021.

\bibitem{liu2023zero}
Ruoshi Liu, Rundi Wu, Basile Van~Hoorick, Pavel Tokmakov, Sergey Zakharov, and
  Carl Vondrick.
\newblock Zero-1-to-3: Zero-shot one image to 3d object.
\newblock {\em CVPR}, 2023.

\bibitem{liu2021semi}
Shaowei Liu, Hanwen Jiang, Jiarui Xu, Sifei Liu, and Xiaolong Wang.
\newblock Semi-supervised 3d hand-object poses estimation with interactions in
  time.
\newblock In {\em CVPR}, 2021.

\bibitem{liu2019soft}
Shichen Liu, Tianye Li, Weikai Chen, and Hao Li.
\newblock Soft rasterizer: A differentiable renderer for image-based 3d
  reasoning.
\newblock In {\em ICCV}, 2019.

\bibitem{hoi4d}
Yunze Liu, Yun Liu, Che Jiang, Kangbo Lyu, Weikang Wan, Hao Shen, Boqiang
  Liang, Zhoujie Fu, He Wang, and Li Yi.
\newblock Hoi4d: A 4d egocentric dataset for category-level human-object
  interaction.
\newblock In {\em CVPR}, 2022.

\bibitem{loshchilov2017decoupled}
Ilya Loshchilov and Frank Hutter.
\newblock Decoupled weight decay regularization.
\newblock {\em ICLR}, 2017.

\bibitem{lu2021omnimatte}
Erika Lu, Forrester Cole, Tali Dekel, Andrew Zisserman, William~T Freeman, and
  Michael Rubinstein.
\newblock Omnimatte: Associating objects and their effects in video.
\newblock In {\em CVPR}, 2021.

\bibitem{melaskyriazi2023realfusion}
Luke Melas-Kyriazi, Christian Rupprecht, Iro Laina, and Andrea Vedaldi.
\newblock Realfusion: 360° reconstruction of any object from a single image.
\newblock In {\em CVPR}, 2023.

\bibitem{mescheder2019occupancy}
Lars Mescheder, Michael Oechsle, Michael Niemeyer, Sebastian Nowozin, and
  Andreas Geiger.
\newblock Occupancy networks: Learning 3d reconstruction in function space.
\newblock In {\em CVPR}, 2019.

\bibitem{mildenhall2021nerf}
Ben Mildenhall, Pratul~P Srinivasan, Matthew Tancik, Jonathan~T Barron, Ravi
  Ramamoorthi, and Ren Ng.
\newblock Nerf: Representing scenes as neural radiance fields for view
  synthesis.
\newblock {\em ECCV}, 2020.

\bibitem{mu2021sdf}
Jiteng Mu, Weichao Qiu, Adam Kortylewski, Alan Yuille, Nuno Vasconcelos, and
  Xiaolong Wang.
\newblock A-sdf: Learning disentangled signed distance functions for
  articulated shape representation.
\newblock In {\em ICCV}, 2021.

\bibitem{newcombe2015dynamicfusion}
Richard~A Newcombe, Dieter Fox, and Steven~M Seitz.
\newblock Dynamicfusion: Reconstruction and tracking of non-rigid scenes in
  real-time.
\newblock In {\em CVPR}, 2015.

\bibitem{nichol2021glide}
Alex Nichol, Prafulla Dhariwal, Aditya Ramesh, Pranav Shyam, Pamela Mishkin,
  Bob McGrew, Ilya Sutskever, and Mark Chen.
\newblock Glide: Towards photorealistic image generation and editing with
  text-guided diffusion models.
\newblock {\em ICML}, 2021.

\bibitem{nie2022pose2room}
Yinyu Nie, Angela Dai, Xiaoguang Han, and Matthias Nie{\ss}ner.
\newblock Pose2room: understanding 3d scenes from human activities.
\newblock In {\em ECCV}, 2022.

\bibitem{oikonomidis2011full}
Iason Oikonomidis, Nikolaos Kyriazis, and Antonis~A Argyros.
\newblock Full dof tracking of a hand interacting with an object by modeling
  occlusions and physical constraints.
\newblock In {\em ICCV}, 2011.

\bibitem{park2019deepsdf}
Jeong~Joon Park, Peter Florence, Julian Straub, Richard Newcombe, and Steven
  Lovegrove.
\newblock Deepsdf: Learning continuous signed distance functions for shape
  representation.
\newblock In {\em CVPR}, 2019.

\bibitem{park2021nerfies}
Keunhong Park, Utkarsh Sinha, Jonathan~T Barron, Sofien Bouaziz, Dan~B Goldman,
  Steven~M Seitz, and Ricardo Martin-Brualla.
\newblock Nerfies: Deformable neural radiance fields.
\newblock In {\em ICCV}, 2021.

\bibitem{park2021hypernerf}
Keunhong Park, Utkarsh Sinha, Peter Hedman, Jonathan~T Barron, Sofien Bouaziz,
  Dan~B Goldman, Ricardo Martin-Brualla, and Steven~M Seitz.
\newblock Hypernerf: A higher-dimensional representation for topologically
  varying neural radiance fields.
\newblock {\em ToG}, 2021.

\bibitem{paszke2019pytorch}
Adam Paszke, Sam Gross, Francisco Massa, Adam Lerer, James Bradbury, Gregory
  Chanan, Trevor Killeen, Zeming Lin, Natalia Gimelshein, Luca Antiga, et~al.
\newblock Pytorch: An imperative style, high-performance deep learning library.
\newblock {\em NeurIPS}, 2019.

\bibitem{Patel2022}
Austin Patel, Andrew Wang, Ilija Radosavovic, and Jitendra Malik.
\newblock Learning to imitate object interactions from internet videos.
\newblock {\em arXiv}, 2022.

\bibitem{petrov2023object}
Ilya~A Petrov, Riccardo Marin, Julian Chibane, and Gerard Pons-Moll.
\newblock Object pop-up: Can we infer 3d objects and their poses from human
  interactions alone?
\newblock In {\em CVPR}, 2023.

\bibitem{poole2022dreamfusion}
Ben Poole, Ajay Jain, Jonathan~T. Barron, and Ben Mildenhall.
\newblock Dreamfusion: Text-to-3d using 2d diffusion.
\newblock {\em ICLR}, 2022.

\bibitem{pumarola2021d}
Albert Pumarola, Enric Corona, Gerard Pons-Moll, and Francesc Moreno-Noguer.
\newblock D-nerf: Neural radiance fields for dynamic scenes.
\newblock In {\em CVPR}, 2021.

\bibitem{radford2021learning}
Alec Radford, Jong~Wook Kim, Chris Hallacy, Aditya Ramesh, Gabriel Goh,
  Sandhini Agarwal, Girish Sastry, Amanda Askell, Pamela Mishkin, Jack Clark,
  et~al.
\newblock Learning transferable visual models from natural language
  supervision.
\newblock In {\em ICLR}, 2021.

\bibitem{Rombach2021HighResolutionIS}
Robin Rombach, A. Blattmann, Dominik Lorenz, Patrick Esser, and Bj{\"o}rn
  Ommer.
\newblock High-resolution image synthesis with latent diffusion models.
\newblock {\em CVPR}, 2021.

\bibitem{rombach2022high}
Robin Rombach, Andreas Blattmann, Dominik Lorenz, Patrick Esser, and Bj{\"o}rn
  Ommer.
\newblock High-resolution image synthesis with latent diffusion models.
\newblock In {\em CVPR}, 2022.

\bibitem{mano}
Javier Romero, Dimitrios Tzionas, and Michael~J. Black.
\newblock Embodied hands: Modeling and capturing hands and bodies together.
\newblock {\em SIGGRAPH Asia}, 2017.

\bibitem{rong2021frankmocap}
Yu Rong, Takaaki Shiratori, and Hanbyul Joo.
\newblock Frankmocap: A monocular 3d whole-body pose estimation system via
  regression and integration.
\newblock In {\em ICCV Workshops}, 2021.

\bibitem{shan2020understanding}
Dandan Shan, Jiaqi Geng, Michelle Shu, and David~F Fouhey.
\newblock Understanding human hands in contact at internet scale.
\newblock In {\em CVPR}, 2020.

\bibitem{cohesive}
Dandan Shan, Richard Higgins, and David Fouhey.
\newblock Cohesiv: Contrastive object and hand embedding segmentation in video.
\newblock {\em NeurIPS}, 2021.

\bibitem{sun2021neural}
Guoxing Sun, Xin Chen, Yizhang Chen, Anqi Pang, Pei Lin, Yuheng Jiang, Lan Xu,
  Jingyi Yu, and Jingya Wang.
\newblock Neural free-viewpoint performance rendering under complex
  human-object interactions.
\newblock In {\em MM}, 2021.

\bibitem{suo2021neuralhumanfvv}
Xin Suo, Yuheng Jiang, Pei Lin, Yingliang Zhang, Minye Wu, Kaiwen Guo, and Lan
  Xu.
\newblock Neuralhumanfvv: Real-time neural volumetric human performance
  rendering using rgb cameras.
\newblock In {\em CVPR}, 2021.

\bibitem{taheri2022goal}
Omid Taheri, Vasileios Choutas, Michael~J Black, and Dimitrios Tzionas.
\newblock Goal: Generating 4d whole-body motion for hand-object grasping.
\newblock In {\em CVPR}, 2022.

\bibitem{taheri2020grab}
Omid Taheri, Nima Ghorbani, Michael~J Black, and Dimitrios Tzionas.
\newblock Grab: A dataset of whole-body human grasping of objects.
\newblock In {\em ECCV}, 2020.

\bibitem{tatarchenko2019single}
Maxim Tatarchenko, Stephan~R Richter, Ren{\'e} Ranftl, Zhuwen Li, Vladlen
  Koltun, and Thomas Brox.
\newblock What do single-view 3d reconstruction networks learn?
\newblock In {\em CVPR}, 2019.

\bibitem{tekin2019h+}
Bugra Tekin, Federica Bogo, and Marc Pollefeys.
\newblock H+ o: Unified egocentric recognition of 3d hand-object poses and
  interactions.
\newblock In {\em CVPR}, 2019.

\bibitem{tulsiani2020implicit}
Shubham Tulsiani, Nilesh Kulkarni, and Abhinav Gupta.
\newblock Implicit mesh reconstruction from unannotated image collections.
\newblock {\em arXiv}, 2020.

\bibitem{tzionas2016capturing}
Dimitrios Tzionas, Luca Ballan, Abhilash Srikantha, Pablo Aponte, Marc
  Pollefeys, and Juergen Gall.
\newblock Capturing hands in action using discriminative salient points and
  physics simulation.
\newblock {\em IJCV}, 2016.

\bibitem{wang2023score}
Haochen Wang, Xiaodan Du, Jiahao Li, Raymond~A Yeh, and Greg Shakhnarovich.
\newblock Score jacobian chaining: Lifting pretrained 2d diffusion models for
  3d generation.
\newblock In {\em CVPR}, 2023.

\bibitem{wang2021neus}
Peng Wang, Lingjie Liu, Yuan Liu, Christian Theobalt, Taku Komura, and Wenping
  Wang.
\newblock Neus: Learning neural implicit surfaces by volume rendering for
  multi-view reconstruction.
\newblock {\em NeurIPS}, 2021.

\bibitem{wang2017binge}
Xiaolong Wang, Rohit Girdhar, and Abhinav Gupta.
\newblock Binge watching: Scaling affordance learning from sitcoms.
\newblock In {\em CVPR}, 2017.

\bibitem{watson2022novel}
Daniel Watson, William Chan, Ricardo Martin-Brualla, Jonathan Ho, Andrea
  Tagliasacchi, and Mohammad Norouzi.
\newblock Novel view synthesis with diffusion models.
\newblock {\em ICLR}, 2022.

\bibitem{wei2022self}
Fangyin Wei, Rohan Chabra, Lingni Ma, Christoph Lassner, Michael Zollh{\"o}fer,
  Szymon Rusinkiewicz, Chris Sweeney, Richard Newcombe, and Mira Slavcheva.
\newblock Self-supervised neural articulated shape and appearance models.
\newblock In {\em CVPR}, 2022.

\bibitem{wen2023bundlesdf}
Bowen Wen, Jonathan Tremblay, Valts Blukis, Stephen Tyree, Thomas M{\"u}ller,
  Alex Evans, Dieter Fox, Jan Kautz, and Stan Birchfield.
\newblock Bundlesdf: Neural 6-dof tracking and 3d reconstruction of unknown
  objects.
\newblock In {\em CVPR}, 2023.

\bibitem{wu2022magicpony}
Shangzhe Wu, Ruining Li, Tomas Jakab, Christian Rupprecht, and Andrea Vedaldi.
\newblock Magicpony: Learning articulated 3d animals in the wild.
\newblock {\em CVPR}, 2023.

\bibitem{yang2022banmo}
Gengshan Yang, Minh Vo, Natalia Neverova, Deva Ramanan, Andrea Vedaldi, and
  Hanbyul Joo.
\newblock Banmo: Building animatable 3d neural models from many casual videos.
\newblock In {\em CVPR}, 2022.

\bibitem{YangCVPR2022OakInk}
Lixin Yang, Kailin Li, Xinyu Zhan, Fei Wu, Anran Xu, Liu Liu, and Cewu Lu.
\newblock {OakInk}: A large-scale knowledge repository for understanding
  hand-object interaction.
\newblock In {\em CVPR}, 2022.

\bibitem{yariv2021volume}
Lior Yariv, Jiatao Gu, Yoni Kasten, and Yaron Lipman.
\newblock Volume rendering of neural implicit surfaces.
\newblock {\em NeurIPS}, 2021.

\bibitem{ye2022scene}
Sifan Ye, Yixing Wang, Jiaman Li, Dennis Park, C~Karen Liu, Huazhe Xu, and
  Jiajun Wu.
\newblock Scene synthesis from human motion.
\newblock In {\em SIGGRAPH Asia}, 2022.

\bibitem{ye2022hand}
Yufei Ye, Abhinav Gupta, and Shubham Tulsiani.
\newblock What's in your hands? 3d reconstruction of generic objects in hands.
\newblock In {\em CVPR}, 2022.

\bibitem{ye2023affordance}
Yufei Ye, Xueting Li, Abhinav Gupta, Shalini De~Mello, Stan Birchfield, Jiaming
  Song, Shubham Tulsiani, and Sifei Liu.
\newblock Affordance diffusion: Synthesizing hand-object interactions.
\newblock In {\em CVPR}, 2023.

\bibitem{zhang2023neuraldome}
Juze Zhang, Haimin Luo, Hongdi Yang, Xinru Xu, Qianyang Wu, Ye Shi, Jingyi Yu,
  Lan Xu, and Jingya Wang.
\newblock Neuraldome: A neural modeling pipeline on multi-view human-object
  interactions.
\newblock In {\em CVPR}, 2023.

\bibitem{zhang2020phosa}
Jason~Y. Zhang, Sam Pepose, Hanbyul Joo, Deva Ramanan, Jitendra Malik, and
  Angjoo Kanazawa.
\newblock Perceiving 3d human-object spatial arrangements from a single image
  in the wild.
\newblock In {\em ECCV}, 2020.

\bibitem{zhang2023adding}
Lvmin Zhang and Maneesh Agrawala.
\newblock Adding conditional control to text-to-image diffusion models.
\newblock {\em arXiv}, 2023.

\bibitem{zheng2021inferring}
Qian Zheng, Weikai Wu, Hanting Pan, Niloy Mitra, Daniel Cohen-Or, and Hui
  Huang.
\newblock Inferring object properties from human interaction and transferring
  them to new motions.
\newblock {\em Computational Visual Media}, 2021.

\bibitem{zhou2023sparsefusion}
Zhizhuo Zhou and Shubham Tulsiani.
\newblock Sparsefusion: Distilling view-conditioned diffusion for 3d
  reconstruction.
\newblock In {\em CVPR}, 2023.

\bibitem{alli2015TheYO}
Berk Çalli, Arjun Singh, Aaron Walsman, Siddhartha~S. Srinivasa, P. Abbeel,
  and Aaron~M. Dollar.
\newblock The ycb object and model set: Towards common benchmarks for
  manipulation research.
\newblock {\em ICRA}, 2015.

\end{thebibliography}
}

\clearpage

\title{Diffusion-Guided Reconstruction of Everyday Hand-Object Interaction Clips
\\Supplementary Materials}

\author{Yufei Ye \qquad Poorvi Hebbar \qquad Abhinav Gupta \qquad Shubham Tulsiani  \\
Carnegie Mellon University  \\
} 

\maketitle
\appendix

\appendix
\ificcvfinal\thispagestyle{empty}\fi


\definecolor{first}{rgb}{1.0, .83, 0.3}
\definecolor{second}{rgb}{1.0, 0.93, 0.7}
\def \third {}

\begin{table*}[t]
\footnotesize
\begin{center}

\caption{{Full ablation results of object reconstruction:} Quantitative results for object reconstruction error using F1@5mm and F1@10mm scores and Chamfer Distance (mm). We compare our method with variants that do not optimize per-frame object poses (Sec.\ref{sec:ablate_oTh}), blend hand and object masks in a hard way (Sec.\ref{sec:ablate_blend}), or do not distill certain geometry modality (Sec.~4.2, Tab.~4)}
\label{tab:ablate_other_obj}

\setlength{\tabcolsep}{2pt}
\resizebox{\linewidth}{!}{
\begin{tabular}{l ccc ccc ccc ccc ccc ccc | ccc}
\toprule

& \multicolumn{3}{c}{Mug} 
& \multicolumn{3}{c}{Bottle}
& \multicolumn{3}{c}{Kettle} 
& \multicolumn{3}{c}{Bowl} 
& \multicolumn{3}{c}{Knife} 
& \multicolumn{3}{c}{ToyCar} 
& \multicolumn{3}{c}{Mean}\\

\cmidrule(r){2-4} \cmidrule(r){5-7} \cmidrule(r){8-10} \cmidrule(r){11-13} \cmidrule(r){14-16}
\cmidrule(r){17-19} \cmidrule(r){20-22}

& $F@5$  & $F@10$ & $CD$  
& $F@5$  & $F@10$ & $CD$  
& $F@5$  & $F@10$ & $CD$  
& $F@5$  & $F@10$ & $CD$  
& $F@5$  & $F@10$ & $CD$  
& $F@5$  & $F@10$ & $CD$  
& $F@5$  & $F@10$ & $CD$  
\\

\midrule
no prior & 0.46 & 0.73 & 1.8 & 0.39 & 0.65 & 2.2 & 0.18 & 0.39 & 9.1 & 0.45 & 0.73 & 1.9 & 0.70 & 0.93 & 0.5 & 0.63 & 0.92 & 0.6 & 0.47 & 0.73 & 2.7 \\
hand prior & 0.48 & 0.77 & 1.4 & 0.37 & 0.66 & 1.6 & 0.30 & 0.60 & 3.4 & 0.38 & 0.63 & 4.2 & 0.09 & 0.24 & 5.8 & 0.70 & 0.97 & 0.4 & 0.39 & 0.65 & 2.8 \\
cat. prior & 0.62 & 0.85 & 1.1 &  0.56 &  {0.95} &  {0.6} &  {0.63} &  {0.94} &  {0.7} & 0.35 & 0.58 & 5.8 & 0.44 & 0.94 & 0.8 & 0.77 & 0.98 & 0.4 & 0.56 & 0.87 & 1.6 \\
wo learning pose &  0.67 & 0.86 &  1.0 & 0.39 & 0.85 & 1.1 & 0.26 & 0.62 & 2.4 &  0.79 &  {0.99} &  {0.3} & 0.58 & 0.95 & 0.7 & 0.82 & 0.99 & 0.3 & 0.59 & 0.88 & 1.0 \\
hard blending & 0.54 & 0.80 & 1.4 & 0.51 & 0.90 & 0.8 & 0.29 & 0.66 & 2.5 & 0.60 & 0.90 & 0.8 & 0.65 & 0.95 & 0.6 &  0.83 & 0.99 &  {0.3} & 0.57 & 0.87 & 1.1 \\
$-$ mask & 0.46 & 0.74 & 1.7 & 0.23 & 0.51 & 2.6 & 0.38 & 0.72 & 2.2 & 0.71 & 0.96 & 0.5 &  {0.83} &  0.98 &  {0.3} & 0.77 &  0.99 & 0.3 & 0.56 & 0.82 & 1.3 \\
$-$ normal & 0.48 & 0.77 & 1.4 & 0.21 & 0.44 & 3.7 & 0.25 & 0.49 & 5.2 & 0.38 & 0.63 & 3.9 & 0.10 & 0.22 & 11.4 & 0.75 & 0.95 & 0.5 & 0.36 & 0.58 & 4.3 \\
$-$ depth &  {0.69} &  {0.93} &  {0.6} &  {0.73} & 0.91 & 0.8 &  0.51 &  0.86 &  1.2 & 0.38 & 0.70 & 2.1 &  0.79 &  {0.98} &  0.4 & 0.82 & 0.98 & 0.3 &  {0.65} &  0.89 &  0.9 \\
Ours & 0.64 &  0.86 & 1.0 & 0.54 &  0.92 &  0.7 & 0.43 & 0.77 & 1.5 &  {0.79} &  0.98 &  0.4 & 0.50 & 0.95 & 0.8 &  {0.83} &  {0.99} &  0.3 &  0.62 &  {0.91} &  {0.8} \\







\bottomrule
\vspace{-3.5em}
\end{tabular}
}

\end{center}

\end{table*}


\definecolor{first}{rgb}{1.0, .83, 0.3}
\definecolor{second}{rgb}{1.0, 0.93, 0.7}
\def \third {}

\begin{table}[t]
\footnotesize
\begin{center}

\caption{{Full ablation results of HOI alignment:} Quantitative results for hand-object alignment using Chamfer distance (mm) in hand frame ($\text{CD}_{\text{h}}$). We compare our method with variants that do not optimize per-frame object poses (Sec.\ref{sec:ablate_oTh}), blend hand and object masks in a hard way (Sec.\ref{sec:ablate_blend}), or do not distill certain geometry modality (Sec.~4.2, Tab.~4).} 
\label{tab:ablate_other_hand}

\setlength{\tabcolsep}{2pt}
\resizebox{0.9\linewidth}{!}{
\begin{tabular}{l c c c c c c | c}
\toprule

& \multicolumn{1}{c}{Mug} 
& \multicolumn{1}{c}{Bottle}
& \multicolumn{1}{c}{Kettle} 
& \multicolumn{1}{c}{Bowl} 
& \multicolumn{1}{c}{Knife} 
& \multicolumn{1}{c}{ToyCar} 
& \multicolumn{1}{c}{Mean}\\



\midrule
no prior & 36.0 & 15.4 & 58.2 &  {75.7} &  {29.5} &  {7.1} &  {37.0} \\
hand prior & 34.5 & 18.3 & 57.5 &  87.5 &  71.7 & 60.6 & 55.0 \\
cat. prior & 23.2 & 75.7 & 54.4 & 158.6 & 164.0 & 34.9 & 85.2 \\
wo opt. obj pose & 21.0 &  14.1 &  {41.8} & 167.1 & 127.1 & 33.2 & 67.4 \\
hard blending & 26.1 & 29.9 & 89.2 & 205.8 & 116.1 & 59.6 & 87.8 \\
 $-$ mask & 36.0 & 28.5 & 60.7 & 504.4 & 97.9 & 41.3 & 128.1 \\
 $-$ normal & 394.9 & 284.1 & 107.9 & 235.5 & 286.0 & 296.6 & 267.5 \\
 $-$ depth &  {14.6} &  {12.7} & 45.5 & 270.6 & 160.6 & 24.0 & 88.0 \\
Ours &  18.1 & 15.3 &  42.2 & 101.8 & 91.6 &  23.3 &  48.7 \\






\bottomrule
\vspace{-3.5em}
\end{tabular}
}

\end{center}

\end{table}

In the supplementary materials, we provide more implementation details and experimental results. We discuss the details of differentiable rendering of the HOI scene representation (Sec.~\ref{sec:render}), network architectures (Sec.~\ref{sec:net_arch}), scored distillation sampling of the pretrained diffusion model (Sec.~\ref{sec:sds}), and initialization details (Sec.~\ref{sec:init}). We also describe how to get 2D segmentation masks from in-the-wild clips (Sec.~\ref{sec:seg}). Then, we show generation by the diffusion model (Sec.~\ref{sec:generation}), full quantitative results reported in the main paper (Sec.~\ref{sec:full_weight}).    Furthermore, we also show supporting evidence that optimizing per-frame object poses (Sec.~\ref{sec:ablate_oTh})  and soft blending (Sec.~\ref{sec:ablate_blend}) are both important for better performance. Lastly, we discuss our failure cases in Sec.~\ref{sec:failure}. 

\section{Implementation Details}
\subsection{Differentiable Rendering (Sec. 3.1)}
\label{sec:render}
Given an HOI scene representation at a certain time $t$ consisting of an implicit field for the object and a mesh for the hand, we use differentiable volumetric renderer~\cite{yariv2021volume} and mesh renderer~\cite{liu2019soft,paszke2019pytorch} to get their masks ($M_o,M_h$) and depth ($D_o,D_h$). In order to supervise them with reprojection loss with respect to the ground truth semantic masks, we blend hand and object masks by their predicted depths to obtain the rendered semantic masks $M \equiv B(M_h, M_o, D_h, D_o)$. 

The soft blending is computed as expected light transported to the cameras, similar to blending two-layer surfaces of
in mesh rendering~\cite{paszke2019pytorch}. More specifically, denote $m_h,d_h,m_o,d_o$ as the value at pixel $(i,j)$, \eg $m_h\equiv M_h[i,j]$.  For any pixel $(i,j)$, the blended value is computed as 
\begin{align}
    m = B(m_h, m_o, d_h, d_o) = \frac{\sum_{k=0,1} w_k l_k}{\sum_{k=0,1} w_k + w_{bg}}
\end{align} 
where subscript $k$ denotes the sorted value of hand and object according to the predicted depth; $l_k$ is the one-hot semantic label (all 0 for background). $w_k$ is the weight computed from depth: 
\begin{align}
    w_k = m_k \exp \frac{z_k - \max_{k,i,j}Z_k[i,j] }{\gamma}, z_k = m_k \frac{d^{\text{far}} - d_k}{d^{\text{far}} - d^{\text{near}}}
\end{align}

We show in Sec.~\ref{sec:ablate_blend} that soft blending (with loss in semantic masks) is important for better results and performs favorably to the alternative (hard blending with ordinal depth loss~\cite{zhang2020phosa,hasson2021towards}). 

\subsection{Network Architectures and Training Details (Sec. 3.1 3.2)}
\label{sec:net_arch}
\paragraph{Implicit field. }
We use Multi-Layer Perceptron (MLPs) to implement the neural implicit surface of the object $\phi$. We borrow the architecture in the original VolSDF~\cite{yariv2021volume} and reduce the network capacity to half as we find it to suffice.  More specifically, we stack four-layer blocks of which each is a linear layer with channel dim $64$ followed by a SoftPlus activation. We apply positional encoding to the queried point $X$ with 6 frequencies. 

\paragraph{Conditional diffusion models. } The backbone of the conditional diffusion model is based on the architecture of the text-to-image inpainting model~\cite{nichol2021glide}. More specifically, it is a 16-layer UNet with cross attentions and skip layers. The text condition along with the diffusion step embedding is passed to the bottleneck of the UNet and is fused with the image feature by cross-attention. The text prompt is encoded as CLIP tokens~\cite{radford2021learning}. 

\paragraph{Details of training diffusion model.} We train the diffusion model with batch size 8, learning rate $1e-4$. We use AdamW~\cite{loshchilov2017decoupled} optimizer with weight decay $0.01$ and train for $500k$ iterations. We use linear noise schedule~\cite{rombach2022high}. 

\paragraph{Details of optimizing HOI scene. } We follow the training setup in a reimplementation~\footnote{\url{https://github.com/ventusff/neurecon}} of the original paper~\cite{yariv2021volume}. We optimize the scene with 1024 rays per step, and set initial learning rate $5e-4$ with exponential learning rate scheduler. We use Adam~\cite{kingma2020method} optimizer and optimize for $50k$ iterations per scene. Within a batch, we bias the sampled pixels from the background, hand, and object region with probability $0.35, 0.35, 0.3$ and linearly interpolate the probability to $0.1, 0.1, 0.8$ in order to spend more effective computation on the object of interest, same as HHOR~\cite{huang2022hhor}. 
In the first 100 warm-up iterations, we turn off SDS and only optimize for the reprojection loss and other regularization terms. This will make the optimization more stable. 

\subsection{Score Distillation Sampling (Sec. 3.3)}
\label{sec:sds}
With the pretrained diffusion model, we follow DreamFusion~\cite{poole2022dreamfusion} to distill the learned prior to the 3D representation. The main idea is to let the diffusion model denoise the corrupted renderings and treats the denoised output as `ground truth'.   More specifically, at each optimization step, we randomly sampled a viewpoint with random rotation from $SO(3)$ and random camera distance. Then, we render the geometry renderings $G_o, G_h$ from the given viewpoint in resolution 64x64. Next, we corrupt the geometry rendering of the object with some noise $G_o^i = \sqrt{\bar \alpha_i}G_o + \sqrt{1-\bar \alpha_i }\epsilon$ ($\bar \alpha$ is the noise scheduling, $\epsilon$ is a gaussian noise) and pass it through the diffusion model along with the geometry rendering of the hand and text prompt. 
\begin{align}
    \hat G_o^i = D_\psi(G_o^i | G_h, C)
\end{align}

We set the classifier-free guidance scale to 4, which is different from the original paper where a small guidance scale cannot converge. It is probably because 2D observations provide stronger cues than text thus leading to  easier convergence.

\subsection{Obtaining hand-object masks for in-the-wild clips.}
\label{sec:seg}
While we provide ground truth segmentation masks to all methods on HOI4D,  
we obtain the segmentation masks by off-the-shelf prediction systems~\cite{shan2020understanding,cheng2021stcn,kirillov2020pointrend} for in-the-wild clips. More specifically, we first use a hand-object interaction detector~\cite{shan2020understanding} to detect the location of the hand and the active object in the first frame. Then, given the detected bounding boxes, we use PointRend~\cite{kirillov2020pointrend} to get the corresponding masks. Next, we pass the masks of interest in the first frame to  a video object segmentation system STCN~\cite{cheng2021stcn} and obtain the tracked masks in every frame. 

To automatically filter out the clips with undesirable segmentation quality, we run the STCN to track forward and backward in time and calculate the Intersection over Union (IoU) between the  initial masks and the masks after tracking back. We use clips with IoU higher than 40\% for both hand and object masks.

\subsection{Initialization with Off-the-Shelf Predictions (Sec. 3.3.)}
\label{sec:init}

We use an off-the-shelf hand reconstruction~\cite{rong2021frankmocap} to estimate initial camera poses $T^t_{c\to h}$, hand shape parameter $\beta$, and hand articulation $\theta_A^t$. The off-the-shelf system predicts per-frame 10-dim hand shape parameters $\beta^t$, 48-dim hand poses $\theta^t$, and a weak perspective camera $s^t,t_x^t, t_y^t$. We take the average of shape parameters across all frames to initialize the hand shape parameter. Among the 48-dim predicted hand pose, we use the 45-dim finger articulation $\theta_A^t$ to initialize hand articulation parameter
while use the remaining 3-dim wrist orientation $\theta_w$ as the rotation component of camera pose $T_{c\to h} ^t$. The translation component is computed by converting the predicted weak-perspective camera to a full-perspective camera (we use a pinhole camera with a focal length of 1  and the principal point at the center of the frame following Zhang \etal~\cite{zhang2020phosa}). This is to handle large perspective effects, which are common in daily  videos of indoor scenes. Given focal length $f$ and principal points $p_x, p_y$, the translation component then becomes $l^t= ((t_x^t - p_x)/s^t, (t_y^t - p_y)/s^t, f/s^t)$. To put them together, the initial camera pose in the hand frame is initialized as: 
\begin{align}
    T_{c\to h} ^t = [R^t | l^t] = [\text{Rot}(\theta_w^t)| \begin{pmatrix} (t_x^t - p_x)/s^t \\ (t_y^t - p_y) / s^t \\ f / s^t \end{pmatrix}]
\end{align}

\begin{figure}[t!]
    \centering
    \includegraphics[width=\linewidth]{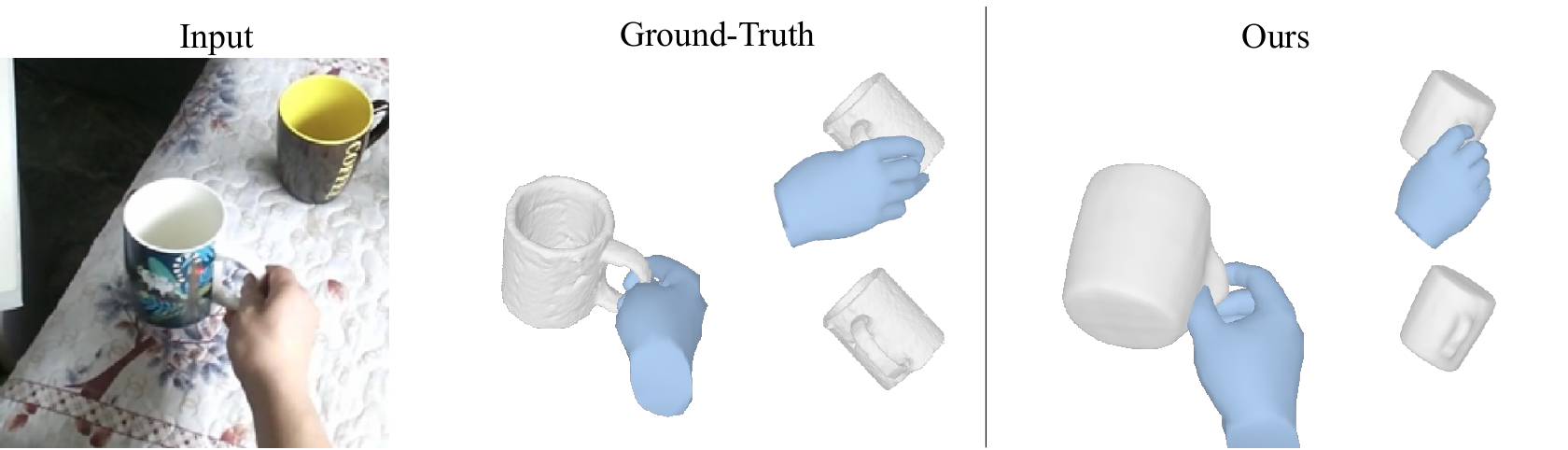}
    \caption{Failure Case}
    \label{fig:failure}
    \vspace{-2mm}
\end{figure}

\begin{figure*}
    \centering
    \includegraphics[width=0.9\linewidth]{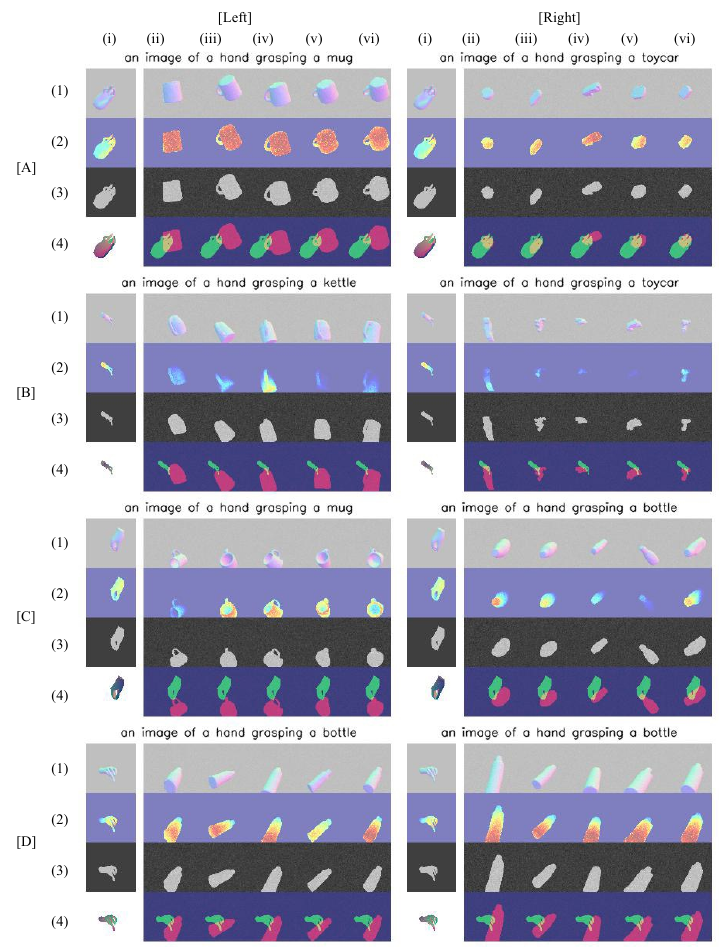}
    \caption{\textbf{Generations from conditional diffusion model.} Given the geometry rendering of hand (i) (row 1-4 visualizing surface normal, depth, mask, and uv coordinate), as well as a text prompt with category information,  we visualize 5 different generations (ii-vi) from the diffusion model. Row 1-3 in col ii-vi shows  the generated geometry rendering of the object, and row 4 visualizes overlayed hand and object masks for a better view of the hand-object relations. All examples on the left use the ground truth paired hand and category information while each example to its right uses another random category but remain hand the same.}
    \label{fig:generation}
\end{figure*}
    
\section{Additional Results}
\subsection{Results of diffusion model generation}
\label{sec:generation}
We show some conditional generations by the pre-trained diffusion model in Fig.~\ref{fig:generation}.  Given the geometry rendering of hand (i) of which row 1-4 visualize surface normal, depth, mask, and uv coordinate, as well as a text prompt with category information,  we visualize 5 different generations (ii-vi) from the diffusion model. Row 1-3 in col ii-vi shows  the generated geometry rendering of the object, and row 4 visualizes overlayed hand and object masks for a better view of the hand-object relations, \ie our model does not output (ii-vi 4). All examples on the left use the ground truth pairs of hand and category information while each example to its right uses another random category but remains hand the same.

As shown in the figure, the generated object matched the category information in the prompt while the generations are diverse in position, orientation, and size. 
Yet, all of the hand-object interactions are realistic, \eg different generated kettle/mug handles all appear at the tip of the hand. 
Comparing left and right examples, different category prompts lead to different generations given the same hand rendering. 
With the same prompt but different hands, the generated objects also change appearance accordingly. For example, in the subfigure [Left A,C], the handles appear at the left when the hand approaches from the left and vice versa. 

Fig.~\ref{fig:generation} indicates that the learned prior is aware of both the hand  prior and the category-level prior hence being informative to guide the 3D reconstruction from clips. 

\subsection{Category-wise results in ablations (Tab. 4)}
\label{sec:full_weight}
In Tab. 4 in the main paper, we only report mean value across all categories due to space limits. We provide quantitative results across all categories in Tab.~\ref{tab:ablate_other_obj} (object reconstruction) and Tab.~\ref{tab:ablate_other_hand} (HOI alignment). 

\subsection{Ablation: Optimizing vs Fixing Object Pose. }
\label{sec:ablate_oTh}
While we observe that the pose of the object in contact relative to hands $T_{h\to o}^t$ does not change much, we still optimize per-frame object poses to account for potential relative motion. As reported in Tab.~\ref{tab:ablate_other_obj}, \ref{tab:ablate_other_hand} and shown on the project page, allowing changing pose across time improves the performance.

\subsection{Ablation: Soft Blending }
\label{sec:ablate_blend}
Our method obtains the final HOI semantic masks by soft blending hand and object rendering as a weighted sum of the labels where the weight depends on their predicted depth. The alternative way is to select the label of the front surface and apply additional ordinal depth loss. This is common in optimizing the interactions of two template meshes~\cite{zhang2020phosa,hasson2021towards}.
As shown in the qualitative results on the webpage, the alternative method generates less desirable hand-object relations as the hand intersects with the object. It is consistent with  quantitative results in Tab.~\ref{tab:ablate_other_obj} and \ref{tab:ablate_other_hand}.

\subsection{Failure Cases}
\label{sec:failure}
 We show one failure case in Fig.~\ref{fig:failure}. The reconstructed mug is in wrong orientation because only semantic masks are used in the reprojection loss. We also struggle with concavity as it is hard to be regularized from only renderings.



\end{document}